\documentclass[runningheads]{llncs}

\PassOptionsToPackage{table}{xcolor}

\usepackage{eccv}

\usepackage{eccvabbrv}

\usepackage{graphicx}
\usepackage{booktabs}

\usepackage{xr-hyper}

\usepackage{hyperref}

\usepackage{orcidlink}

\usepackage{url}
\usepackage{colortbl}
\usepackage{multirow}
\usepackage{subcaption}
\usepackage{amssymb}
\usepackage{pifont}
\usepackage{array}
\usepackage{cellspace}
\usepackage[utf8]{inputenc}
\usepackage{algorithm}
\usepackage{algpseudocode}
\usepackage{amsmath}
\usepackage{mwe}
\usepackage{wrapfig}
\usepackage{tabularx}
\newcolumntype{Y}{>{\raggedright\arraybackslash}X}
\usepackage{enumitem}

\usepackage{etoc}
\usepackage{kotex}

\newcommand{\authcount}[1]{}
\etocsetlevel{title}{6}
\etocsetlevel{author}{6}
\definecolor{mygreen}{RGB}{0, 128, 0}
\newcommand{\good}[1]{\textcolor{mygreen}{\scriptsize{(#1)}}} 
\newcommand{\bad}[1]{\textcolor{red}{\scriptsize{(#1)}}}      
\newcommand{\undund}{\textit{und} $\rightarrow$ \textit{und}}
\newcommand{\undgen}{\textit{und} $\rightarrow$ \textit{gen}}
\newcommand{\genund}{\textit{gen} $\rightarrow$ \textit{und}}
\newcommand{\gengen}{\textit{gen} $\rightarrow$ \textit{gen}}

\begin{document}
\etocdepthtag.toc{main}
 

\title{\textbf{Transferability Between Understanding and Generation in Unified Multimodal Models}
} 

\titlerunning{Transferability in Unified Multimodal Models}

\author{Jiwon Kang\inst{1}\textsuperscript{*} \and
Heeji Yoon\inst{1}\textsuperscript{*} \and
Jaewoo Jung\inst{1} \and
Jaewon Min\inst{1} \and
Minkyeong Jeon\inst{1} \and
Biyeon Hwang\inst{2} \and
Sangwon Jung\inst{3}\textsuperscript{\dag} \and
Seungryong Kim\inst{1}\textsuperscript{\dag}}

\authorrunning{Kang et al.}

\institute{
  $^{1}$\,KAIST AI, South Korea \quad
  $^{2}$\,Blynx, South Korea \quad
  $^{3}$\,Trillionlabs, South Korea
}


\maketitle

\renewcommand{\thefootnote}{}
\footnotetext{
  \textsuperscript{*}~Equal contribution. \\
  \textsuperscript{\dag}~Co-corresponding author.}

\begin{abstract}
Unified Multimodal Models (UMMs) integrate image understanding and generation within a single architecture, yet how the two tasks interact remains understudied. We investigate \textbf{transferability} in UMMs: whether training a capability on one task improves the same capability on the other without explicit supervision. Through controlled experiments, we empirically find that transferability depends on architecture—models with fully shared transformer backbone and a unified visual encoder exhibit consistent cross-task transfer, while loosely coupled designs show little or none. Leveraging this transferability, we propose a practical training strategy. The most straightforward way to improve a target generative capability (\eg, counting) is to fine-tune generation directly, but this can degrade visual quality due to distribution shift. Instead, we train the corresponding understanding task and let it \textit{transfer} into generation, which improves capability-specific generative performance while minimizing distribution shift. We validate this across three capabilities—counting, spatial relation, and text recognition/generation—showing that cross-task transferability can be systematically exploited in UMMs.

\end{abstract}

\section{Introduction}


Propelled by the success of Large Language Models (LLMs)~\cite{openai2023gpt4,claude3,geminiteam2023gemini,touvron2023llama}, Vision-Language Models (VLMs)~\cite{bai2025qwen3, chen2024internvl, guo2025deepseek, liu2023visual, deitke2025molmo, yoon2025visual} have achieved significant breakthroughs in image understanding. Concurrently, diffusion-based generative models~\cite{rombach2022high,podell2024sdxl,bfl2025representation} have driven tremendous advancements in image generation. Despite their profound progress, these two domains have mostly evolved in isolation. This historical separation has recently motivated the development of Unified Multimodal Models (UMMs)~\cite{deng2025bagel, cui2025emu3,chen2505blip3,chen2025januspro,xin2025lumina, xu2025qwen3omni}, which integrate both tasks---image understanding and generation---within a single architecture. While recent UMMs demonstrate that such unification can achieve performance comparable to specialized experts for each task, it remains fundamentally unclear how the two tasks interact once they are trained within a single model.


Several recent studies~\cite{wu2025harmonizing,li2025dual,xu2025pisces,dong2024dreamllm} have investigated interactions between image understanding and generation, arguing that understanding can improve generation and vice versa. However, existing analyses typically examine these interactions through aggregate performance changes on general benchmarks~\cite{li2023pope,yue2024mmmu,fu2023mme,ghosh2023geneval} (\eg, adding generation training improves understanding benchmark scores). While such results provide useful high-level observations, relying on broad benchmark improvements makes it difficult to determine whether the observed gains stem from genuine cross-task knowledge transfer or merely from indirect factors such as additional data or regularization effects.

We address this gap by studying a more direct and actionable signal of cross-task interaction: \emph{transferability}. Instead of inferring interaction from aggregate benchmark gains, we examine whether training a given capability (\eg, counting) in one task (understanding or generation) reliably improves the same capability in the other task, under controlled interventions. Through a series of experiments, we find that such cross-task transferability emerges in current UMMs, and this study leads to two key findings: (i) the existence and strength of transferability vary across models, empirically revealing a dependency on architectural design; (ii) when transfer is present, it can be exploited as a practical mechanism to improve targeted generative capabilities by training on the corresponding understanding task.

Specifically, we begin with a controlled analysis on architecture using \textit{counting} as a probe task. Across several representative open-source UMMs~\cite{xin2025lumina,chen2025januspro,chen2505blip3,deng2025bagel}, we empirically observe that transferability depends on architectural design. In particular, models that employ a fully shared transformer backbone along with a unified visual encoder~\cite{xin2025lumina} exhibit consistently the strongest transfer between tasks, whereas architectures that leverage a separate backbone or an image encoder~\cite{chen2025januspro, chen2505blip3} show little or no transfer. These results suggest that the degree of representation sharing may play an important role in enabling cross-task capability transfer. 

Using the architecture with the strongest transfer, our next step is to investigate whether transferability can be utilized as a practical mechanism for enhancing generative capabilities. While directly fine-tuning with a generation objective may be the most straightforward approach to acquire a specific capability, it carries the risk of inducing a distribution shift~\cite{lv2024pick,lee2024direct,chae2025apt} that may degrade overall visual quality. Instead, we find that training on the corresponding understanding task can effectively introduce the capability into generation through transfer, minimizing the distribution shift. Extensive experiments across multiple visual capabilities, including counting, spatial relation, and text recognition/generation, confirm that cross-task transfer can consistently improve generation performance on specific capabilities while preserving general multimodal understanding ability and generative quality. 


Our contributions are summarized as follows:
\begin{itemize}
    \item We introduce \textbf{transferability} as a tool for analyzing cross-task interactions in Unified Multimodal Models, and show that capabilities learned in understanding or generation can be transferred to the other task.
    \item Through controlled experiments, we empirically demonstrate that transferability depends on architectural design, emerging most clearly in models with fully shared transformer backbones with a unified visual encoder.
    \item We show that transferability can be leveraged as a practical strategy for improving targeted generative capabilities while avoiding degradation of original multimodal understanding ability or generative quality, and validate this approach across counting, spatial relation, and text recognition/generation tasks.
\end{itemize}

\section{Related Work}
\subsection{Image Understanding and Generation in UMMs}
\label{subsec:taxonomy}
The success of large language models~\cite{openai2023gpt4, touvron2023llama,claude3,geminiteam2023gemini} has catalyzed the extension of language modeling into the multimodal domain. Vision-Language Models (VLMs)~\cite{bai2025qwen3, chen2024internvl, guo2025deepseek, liu2023visual, deitke2025molmo} enable strong image understanding by pairing a visual encoder with an LLM. More recently, Unified Multimodal Models (UMMs) have pushed this paradigm further by incorporating image generation alongside understanding within a single model~\cite{team2024chameleon, wu2025janus, deng2025bagel, xin2025lumina, chen2505blip3}.

\vspace{-10pt}
\subsubsection{Visual representations.}
A key design axis in UMMs is the choice of visual representation for encoding input images. One line of work adopts language-supervised encoders such as CLIP~\cite{radford2021clip} and SigLIP~\cite{zhai2023sigmoid,tschannen2025siglip}, whose representations are aligned with text and thus preserve high-level semantics amenable to reasoning. A complementary line leverages autoencoders~\cite{ballard1987modular} which are typically employed for image generation. These autoencoders, which prioritize pixel-level reconstruction fidelity, are usually trained with variational objectives~\cite{kingma2013vae} for continuous representation~\cite{rombach2022high, podell2024sdxl, esser2024scaling,labs2025flux,bfl2025representation} or a vector quantization objective~\cite{van2017neural} for discrete representation~\cite{van2017neural, esser2021taming}.
Analyses in Janus~\cite{chen2025januspro,wu2025janus} and related works~\cite{tong2024cambrian} show that language-supervised representations substantially outperform compression-oriented tokenizers for understanding tasks, motivating several UMMs to employ separate encoders for each task.

\vspace{-10pt}
\subsubsection{Generation objectives.}
Image generation in UMMs follows two main paradigms. \emph{Autoregressive} models~\cite{wu2025janus,chen2025januspro, team2024chameleon, cui2025emu3,kou2025orthus,wu2025harmonizing} employ discrete image tokenizers and generate image tokens sequentially, directly extending the next-token prediction framework of LLMs. \emph{Diffusion}-based models instead learn a denoising objective~\cite{song2019generative, ho2020denoising, song2020denoising, lipman2022flow, liu2022flow, albergo2022building} and can be further divided into discrete-token diffusion models~\cite{yang2025mmada, xin2025lumina, shi2025muddit} which apply discrete diffusion~\cite{austin2021structured,lou2023discrete,sahoo2024simple} over quantized latent representations, and continuous diffusion models~\cite{xie2025show,zhou2024transfusion, chen2505blip3,dong2024dreamllm, sun2024emu,ge2024seed} which operate on continuous latent representations.

\vspace{-10pt}
\subsubsection{Architectural taxonomy of unified multimodal models.}
Although finer-grained categorizations are possible~\cite{zhao2025unified}, we group existing unified models into three dominant architecture families for our analysis. \textbf{(1) Single transformer} designs route all modalities through a shared transformer backbone, maximizing parameter sharing~\cite{yang2025mmada,team2024chameleon, cui2025emu3,wang2024emu3,xin2025lumina,wu2025janus,chen2025januspro,shi2025muddit}. These models either use unified image encoders~\cite{team2024chameleon, yang2025mmada,xin2025lumina,wang2024emu3,shi2025muddit} or separate ones~\cite{wu2025janus,chen2025januspro} for understanding/generation. 
\textbf{(2) Mixture-of-Transformers (MoT)} designs maintain modality-specific transformer branches but couple them through joint attention layers, allowing selective parameter sharing while retaining specialized capacity~\cite{deng2025bagel,li2025unifork,wang2025lightfusion, wang2025hbridge}.
\textbf{(3) Decoupled transformer} designs employ separate transformers for language and vision, where the LLM provides conditioning signals (\eg, hidden states) that are subsequently fed into a dedicated diffusion module for generation~\cite{pan2025meataquery,tong2025metamorph,chen2505blip3,sun2024emu,ge2024seed}.

\subsection{Understanding Interactions Between Understanding and Generation in UMMs}
A central question in designing Unified Multimodal Models is whether multimodal understanding and generation genuinely benefit each other within a shared architecture. Several works report potential synergy, arguing that high-level semantic representations learned for understanding can improve the semantic fidelity of generation, while fine-grained visual representations acquired through generation may enhance visual grounding and compositional reasoning~\cite{wu2026liquid, xin2025lumina, yang2025mmar, wu2025harmonizing, kou2025orthus, li2025dual,zhang2025unified}. On the other hand, research such as UniFork~\cite{li2025unifork} empirically demonstrates that ideal representations for these two tasks throughout the network layers are distinct, suggesting that simple parameter sharing can lead to a representational compromise. This perspective aligns with a broader trend of decoupled architectures that utilize task-specific branches in higher layers or separated visual paths to mitigate potential conflicts~\cite{wu2025janus, chen2025januspro, deng2025bagel}. 

While these studies provide valuable insights into the potential benefits and trade-offs of joint training, most analyses examine interaction through aggregate performance changes on general benchmarks. For example, prior work often evaluates whether increasing generation-oriented training improves overall understanding benchmarks (\eg, MMMU~\cite{yue2024mmmu} or POPE~\cite{li2023pope}), or conversely whether additional understanding supervision improves generation benchmarks such as GenEval. Although such evaluations provide useful high-level signals, they make it difficult to isolate how specific capabilities learned in one modality influence the other. In contrast, our work investigates cross-task interaction at the capability level through the lens of transferability. Rather than inferring interaction from aggregate benchmark improvements, we directly examine whether learning a capability in one modality improves the same capability in the other modality. 
\section{Does knowledge transfer emerge in UMMs?}

\begin{figure*}[!t]
    \centering
    \includegraphics[width=\linewidth]{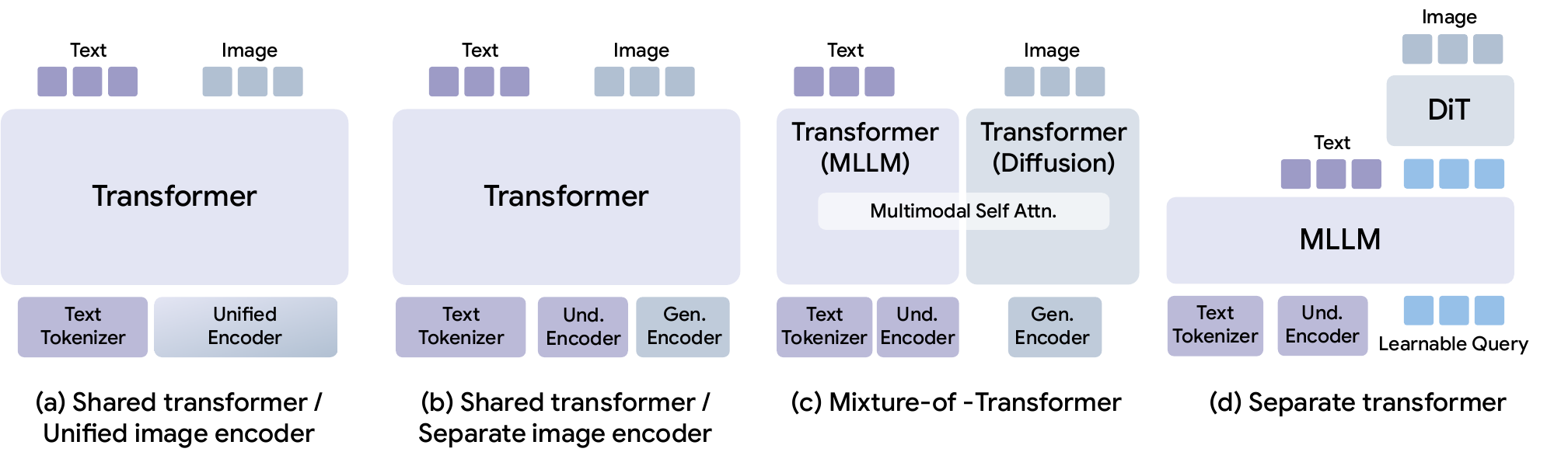} 
    \vspace{-5pt}
    \caption{\textbf{Architecture of chosen unified multimodal models for cross-task transferability analysis on counting.} 
    (a) Shared Transformer with a unified image encoder.
    (b) Shared Transformer with separate image encoders for each task (understanding/generation).
    (c) Mixture-of-Transformer which uses separate transformers for understanding and generation but connects each other via joint(multimodal) self-attention. 
    (d) Separate Transformer where the MLLM transformer provides conditioning signals that are fed into a diffusion transformer (DiT).}
    \label{fig:arch_categorize}
\end{figure*}

\begin{table*}[!t]
\centering
\resizebox{\textwidth}{!}{%
\begin{tabular}{c||cc|cc|cc|cc}
\toprule
\multirow{2}{*}{\textbf{Train $\rightarrow$ Test}} 
& \multicolumn{2}{c|}{\textbf{(a) Lumina-DiMOO}} 
& \multicolumn{2}{c|}{\textbf{(b) Janus-Pro}} 
& \multicolumn{2}{c|}{\textbf{(c) BAGEL}} 
& \multicolumn{2}{c}{\textbf{(d) BLIP3-o}} \\
& Acc. (\%)~$\uparrow$ & MAD~$\downarrow$
& Acc. (\%)~$\uparrow$ & MAD~$\downarrow$
& Acc. (\%)~$\uparrow$ & MAD~$\downarrow$
& Acc. (\%)~$\uparrow$ & MAD~$\downarrow$ \\

\midrule

\rowcolor{gray!15}
\multicolumn{9}{l}{\textit{Generation Evaluation}} \\
Baseline
& 48.0 & 1.11 & 38.0 & 1.46 & 42.0 & 1.16 & 31.0 & 2.08 \\
\undgen& 
57.0 \textbf{\good{+9.0}} & 0.83 \textbf{\good{-0.28}} &  
32.0 \bad{-6.0} & 1.55 \bad{+0.09} & 
47.0 \good{+5.0} & 1.11 \good{-0.05} & 
37.0 \good{+6.0} & 1.80 \good{-0.28} \\

\midrule

\rowcolor{gray!15}
\multicolumn{9}{l}{\textit{Understanding Evaluation}} \\
Baseline & 
22.0 & 2.48 & 
37.0 & 2.70 & 
63.0 & 0.65 & 
63.0 & 0.54 \\

\genund&
30.0 \textbf{\good{+8.0}} & 1.88 \textbf{\good{-0.60}} & 
38.0 \good{+1.0} & 3.34 \bad{+0.64} & 
64.0 \good{+1.0} & 0.62 \good{-0.03} & 
62.0 \bad{-1.0} & 0.56 \bad{+0.02} \\

\bottomrule

\end{tabular}
}
\vspace{10pt}
\caption{
    \textbf{Probing cross-task transferability on counting across different UMM architectures.} We conduct two separate fine-tuning experiments and compare them against the baseline. (i) \undgen: training only on understanding and evaluating on generation. (ii) \genund: training only on generation and evaluating on understanding. Acc.\ denotes exact-match accuracy (\%) and MAD denotes mean absolute deviation from the ground-truth count. Shared transformer models with a unified visual tokenizer (Lumina-DiMOO) show the strongest transferability in both directions.
}
\label{tab:transfer}
\vspace{-25pt}
\end{table*}

\label{sec:umms}


 In this section, we investigate cross-task transfer in current open-source UMMs: (i) whether learning a capability in one task (understanding or generation) improves the same capability in the other, and (ii) whether this transfer depends on architectural design. To this end, we conduct an analysis using counting as a probe capability, which provides a clean and quantifiable concept that both understanding and generation should encode. 

\vspace{-10pt}
\subsection{Experimental setup} 

Considering the diverse architectural choices introduced in Section~\ref{subsec:taxonomy}, we select four representative open-source models: Lumina-DiMOO~\cite{xin2025lumina} for \textbf{(a) Shared Transformer with unified image encoder}, Janus-Pro~\cite{chen2025januspro} for \textbf{(b) Shared Transformer with separate image encoder}, BAGEL~\cite{deng2025bagel} for \textbf{(c) Mixture of Transformers}, and BLIP3-o~\cite{chen2505blip3} for \textbf{(d) Separate Transformer}. The architectural overview of each selected model is illustrated in Figure~\ref{fig:arch_categorize}. We select 7B--8B parameter versions of each model.


For each model, we conduct two separate training runs (understanding and generation) using LoRA-based supervised fine-tuning. We construct training datasets by filtering PixMo-Points and PixMo-Count~\cite{deitke2025molmo} to a counting range of 0--20. For understanding, the model learns counting through visual question answering, where it answers how many objects appear in an image. 
For generation, the model learns counting through text-to-image generation, where it is supervised to produce images containing objects whose count matches the prompt. To get the generation prompts, we use Qwen3-VL~\cite{bai2025qwen3} to caption the same images and append object count information to form the generation prompts. 


\begin{figure*}[t]
    \centering
    \includegraphics[width=\linewidth]{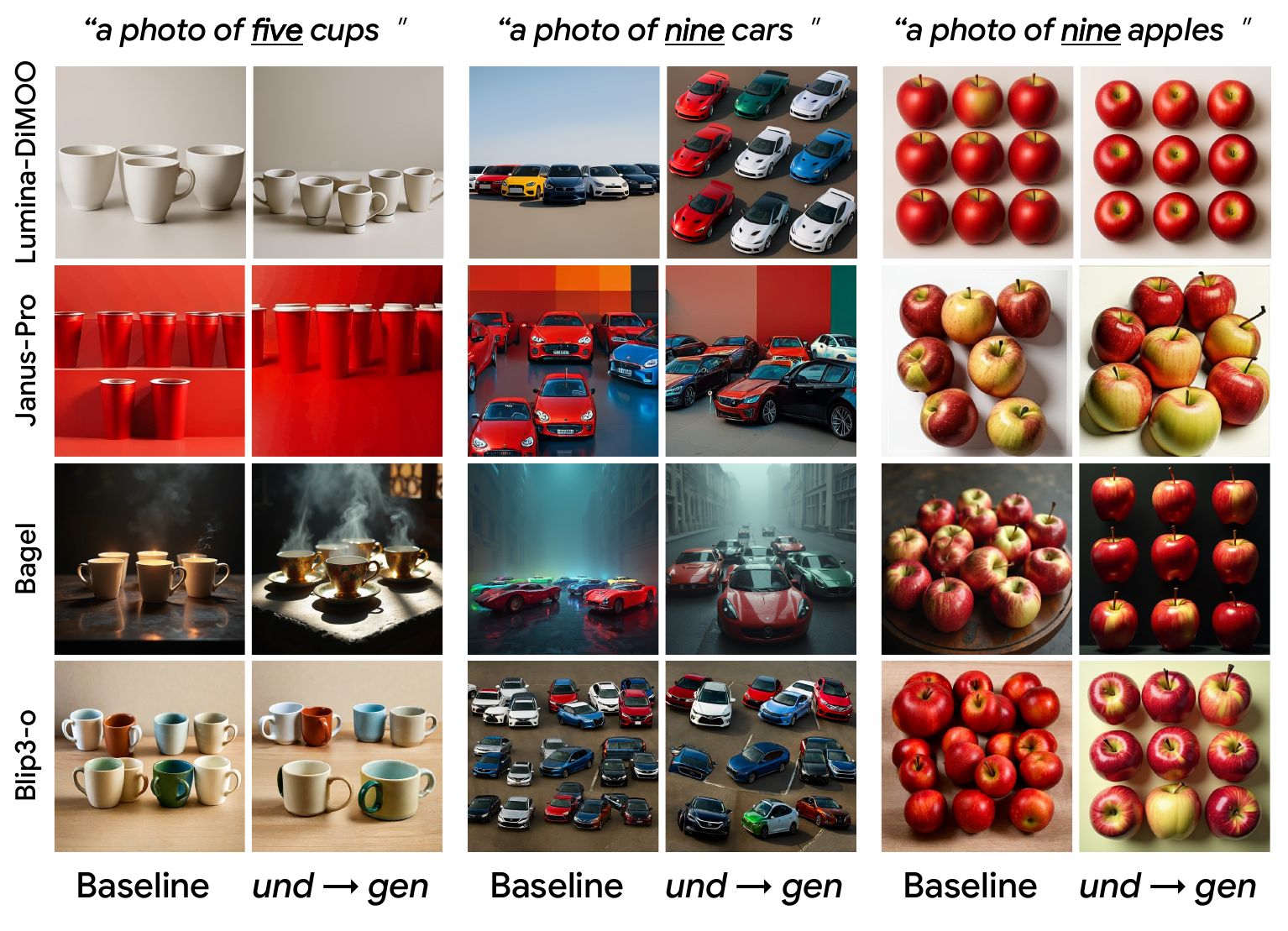} 
    \vspace{-20pt}
    \caption{\textbf{Qualitative results of \undgen\ transferability across models.} 
    For each prompt, we compare the baseline model with the model trained only on counting understanding data (\undgen) and evaluated on generation. All images are generated using the same random seed for fair comparison.}
    \label{fig:counting_quals}
    \vspace{-15pt}
\end{figure*}

For evaluation, we use two separate protocols. Understanding is evaluated on the test split of PixMo-Count~\cite{deitke2025molmo} in a VQA format, where the model is asked to count target objects in an image and responds with a number. For generation, we follow the counting protocol from GenEval~\cite{ghosh2023geneval} and construct prompts in the form \texttt{"A photo of [NUMBER] [OBJECT]s"} with counts ranging from 2 to 10. Generated images are evaluated using an object detector~\cite{cheng2022masked} to verify whether the number of detected objects matches the specified count. We report both accuracy and mean absolute deviation (MAD) to capture exact correctness and proximity to the ground truth.

\vspace{-10pt}
\subsection{\textbf{Transferability across architectures}}
\subsubsection{Transferability varies across architectures.}
Table~\ref{tab:transfer} summarizes the results. The most consistent finding is that knowledge transfer from understanding to generation, which we term \textit{understanding-to-generation transfer} (\undgen), is observed in all models except Janus-Pro~\cite{chen2025januspro}. BAGEL~\cite{deng2025bagel}, Lumina-DiMOO~\cite{xin2025lumina}, and \mbox{BLIP3-o~\cite{chen2505blip3}} all show improved generation accuracy and lower MAD when trained with the understanding objective, indicating that counting capability acquired through understanding can be transferred to generation. In contrast, the inverse direction (\genund) is more selective: only BAGEL and Lumina-DiMOO show gains in counting understanding from generation training, making transfer bi-directional in these two models. Among all models, Lumina-DiMOO shows the strongest effect in both directions, underscoring that transferability is not uniform across UMM architectures. Qualitative examples illustrating this behavior are shown in Figure~\ref{fig:counting_quals}.

\vspace{-10pt}
\subsubsection{Which architecture results in the strongest transferability?} From an architectural perspective, we empirically observe that models which employ a fully shared transformer backbone for both understanding and generation, along with a unified vision encoder (Figure~\ref{fig:arch_categorize}-(a)), such as Lumina-DiMOO~\cite{xin2025lumina}, exhibit the strongest transferability. This result is consistent with recent empirical findings that adopting a unified encoder for image understanding and generation produces synergistic effects in UMMs~\cite{wu2025harmonizing, wu2026liquid}: because a shared transformer backbone with a unified encoder lets both tasks operate over a shared representation, knowledge acquired through one objective naturally propagates to the other, facilitating cross-task transfer. By contrast, architectures that maintain separate visual pathways for each task offer fewer opportunities for such propagation, which explains the weaker or absent transfer observed in models like Janus-Pro.

\section{Using Transferability to Improve Generative Capabilities}
\label{sec:tasks}

In this section, we investigate how the transferability that emerged in UMMs can be practically exploited. In practice, generative models often exhibit weaknesses in specific capabilities such as spatial relation~\cite{liu2022compositional, feng2022training, gokhale2022visor, chen2024training, phung2024grounded, xiao2024r,wang2026everything} or object counting~\cite{fu2025counting,boo2025countsteer,binyamin2025make,li2025countdiffusion}. Generation objectives~\cite{song2019generative, ho2020denoising, song2020denoising, lipman2022flow, liu2022flow, albergo2022building} fundamentally learn to approximate their training data distribution, so a straightforward way to inject a specific skill (\eg, counting) is fine-tuning directly on a skill-focused dataset with generation objectives. However, fine-tuning pulls the model away from its original image generation distribution, which can cause a distribution shift that may degrade generative quality~\cite{lv2024pick,lee2024direct,chae2025apt}.

This issue is even more pronounced for recent models whose generation distribution is carefully aligned with human preferences in a final post-training stage, using SFT over curated high-quality datasets~\cite{podell2024sdxl,dai2023emu} or RL-based methods~\cite{fan2023dpok, lee2023aligning,yang2024using, li2024aligning, wallace2024diffusion, black2024training}. However, direct fine-tuning readily destroys this alignment. Workarounds would be either curating a dataset that is simultaneously high-quality and skill-rich, or repeating the costly post-training alignment after fine-tuning, neither of which is practical. 


We propose a practical alternative that leverages transferability. Rather than optimizing the generation objective directly, we train the target capability through the understanding and let it transfer to generation. This approach enables targeted improvements in generative capabilities while maintaining the overall generation quality of the model. To study this mechanism, we analyze understanding-to-generation transfer across three representative tasks, \ie, counting, spatial relation, and text recognition/generation. We conduct the following experiments on Lumina-DiMOO~\cite{xin2025lumina}, as its single-transformer design with a unified image encoder empirically exhibited the strongest cross-task transfer among the UMMs studied in Section~\ref{sec:umms}. Result on another model~\cite{yang2025mmada} sharing the same architecture is reported in the Appendix.

\begin{figure*}[t]
    \centering
    \includegraphics[width=\linewidth]{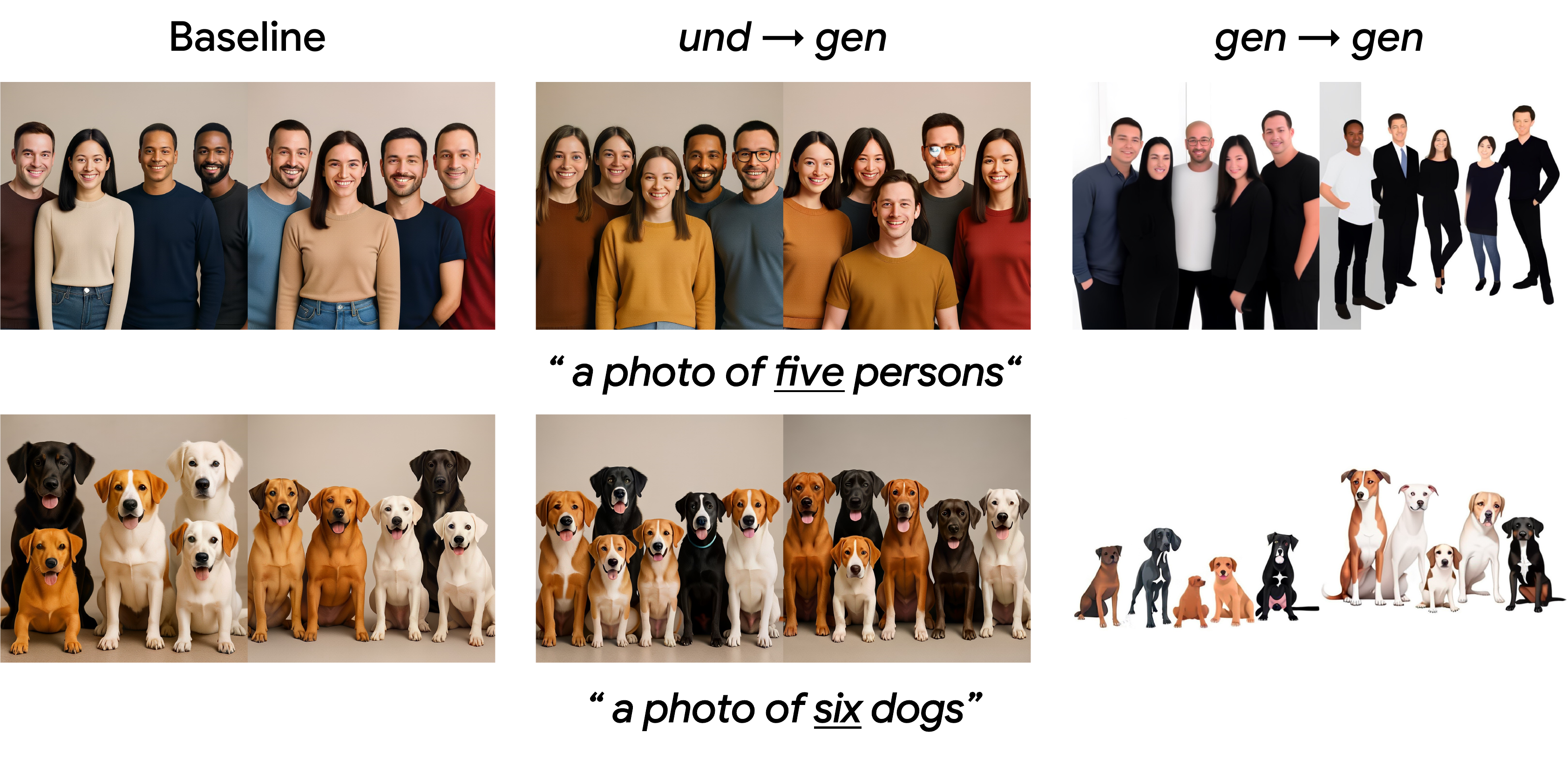} 
    \vspace{-20pt}
    \caption{\textbf{Qualitative comparison of understanding-to-generation transfer (\undgen) versus direct generation training (\gengen) on counting.} While direct generation training (\gengen) significantly degrades image quality due to distribution shift, \undgen\ preserves generation quality while achieving improved counting accuracy.}
    \label{fig:undgen_counting}
    \vspace{-5pt}
\end{figure*}



\begin{table}[t]
    \centering
    \setlength{\tabcolsep}{6pt}
    \renewcommand{\arraystretch}{0.85}
    \begin{tabular}{c|cccc}
        \toprule
        \textbf{Train $\rightarrow$ Test}
        & \textbf{Acc. (\%)}~$\uparrow$
        & \textbf{MAD}~$\downarrow$
        & \textbf{IS}~\cite{salimans2016isscore}~$\uparrow$
        & \textbf{FID}~\cite{heusel2017fid}~$\downarrow$
        \\
        \midrule
        Baseline
        & 48.0 & 1.11 & 16.86 & - \\
        \undgen
        & \textbf{57.0} \good{+9.0} & \textbf{0.83} \good{-0.28} & \textbf{17.55}  & \textbf{31.47} \\
        \gengen
        & 57.0 \good{+9.0} & 0.91 \good{-0.20} & 15.29  & 52.51  \\
        \bottomrule
    \end{tabular}
    \vspace{5pt}

    \caption{\textbf{Quantitative comparison of understanding-to-generation transfer (\undgen) versus direct generation training (\gengen) on counting.} Using transferability (\undgen) attains task accuracy on par with direct training (\gengen) while retaining image quality (IS~\cite{salimans2016isscore}, FID~\cite{heusel2017fid}). In contrast, \gengen\ degrades both metrics.}
    \vspace{-25pt}
    
    \label{tab:gen_quality}
\end{table}

\subsection{Counting}

We first revisit the counting generation from this perspective. Table~\ref{tab:transfer} previously showed that counting exhibits strong transferability from understanding to generation. To examine whether this transfer can be practically beneficial, we compare the model trained with the understanding objective (\undgen) with a model directly trained with the generation objective (\gengen) in Table~\ref{tab:gen_quality}. Surprisingly, the model trained through understanding achieves a lower mean absolute deviation (MAD) in counting generation. In other words, introducing counting capability through the understanding leads to more accurate counting behavior in generation than directly optimizing the generation objective. 

The qualitative results in Figure~\ref{fig:undgen_counting} further highlight this difference. When the model is trained with the generation objective (\gengen), the narrow distribution of the dataset causes the model to drift from its original training distribution, resulting in noticeable degradation of overall image quality. In contrast, the model trained via understanding (\undgen) preserves the overall image quality. To quantify this, we additionally measure distribution shift using Fréchet Inception Distance (FID)~\cite{heusel2017fid} and image quality using Inception Score (IS)~\cite{salimans2016isscore} in Table~\ref{tab:gen_quality}. Specifically, FID is computed between the baseline generated images and those from the trained models. Training directly with the generation objective (\gengen) yields a higher FID than \undgen\ and degrades IS relative to the baseline, confirming both a larger distribution shift and loss of generative quality. In contrast, the understanding-based approach (\undgen) maintains generation quality with lower distribution shift, while improving counting accuracy, demonstrating that transfer through understanding provides a more stable way to introduce new capabilities into generation.

\subsection{Spatial relation}
\label{sec:spatial_rel}
We next consider a structurally different capability, spatial relation, where the model must encode the relative layout of objects. Unlike counting, this task requires capturing compositional relationships between multiple entities. This allows us to examine whether the transferability observed in counting extends to more complex relational concepts.

\begin{wrapfigure}[12]{r}{0.4\linewidth}
    \centering
    \vspace{-30pt}
    \includegraphics[width=\linewidth]{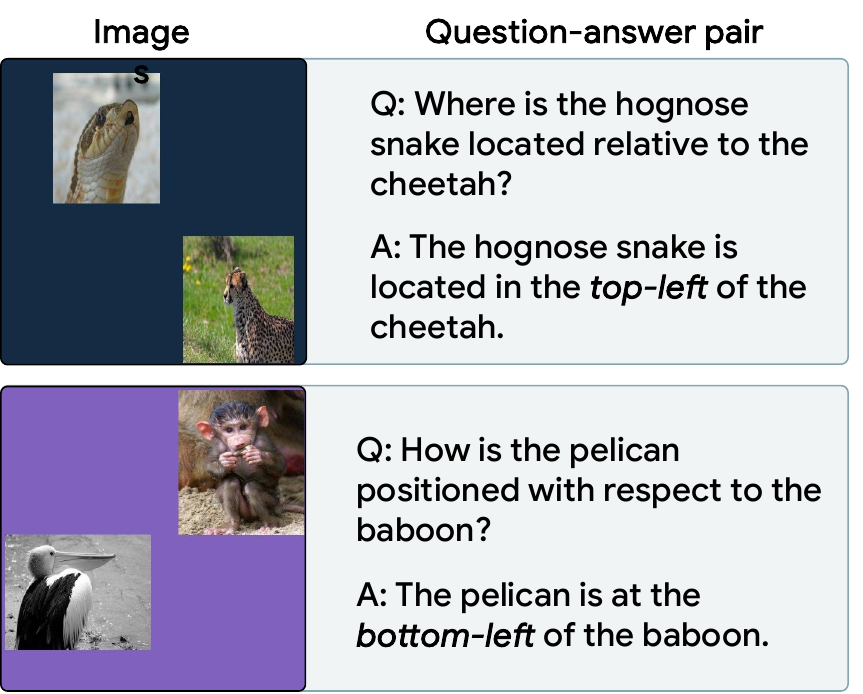}
    \vspace{-15pt}
    \caption{\textbf{Example of synthetic dataset for spatial relation.} }
    \label{fig:spatial_dataset}
    \vspace{0pt}
    \end{wrapfigure}

\subsubsection{Experimental setting.} We observe that pretrained models already handle simple spatial relations such as left, right, top, and bottom reasonably well. We therefore focus on a more challenging setting: diagonal relations between two objects (\eg, top-left, top-right, bottom-right, bottom-left). Since existing spatial reasoning datasets~\cite{hudson2019gqa, zhang2021vsr, fan2025gritteachingmllmsthink,blip3-xgenmm,deng2025internspatial,yang2019spatialsense,song2025robospatial} are difficult to use directly for our purpose---owing to their limited scale, mismatched spatial complexity, or inconsistent annotation formats---we construct a synthetic dataset of 200K samples from ImageNet~\cite{deng2009imagenet}. For each sample, as shown in Figure~\ref{fig:spatial_dataset} we select two class–image pairs and place them on a canvas according to a randomly sampled diagonal direction. Each sample is paired with a prompt of the form \texttt{"[class A] is located in the [direction] of [class B]"}.

\begin{figure*}[t]
    \centering
    \includegraphics[width=\linewidth]{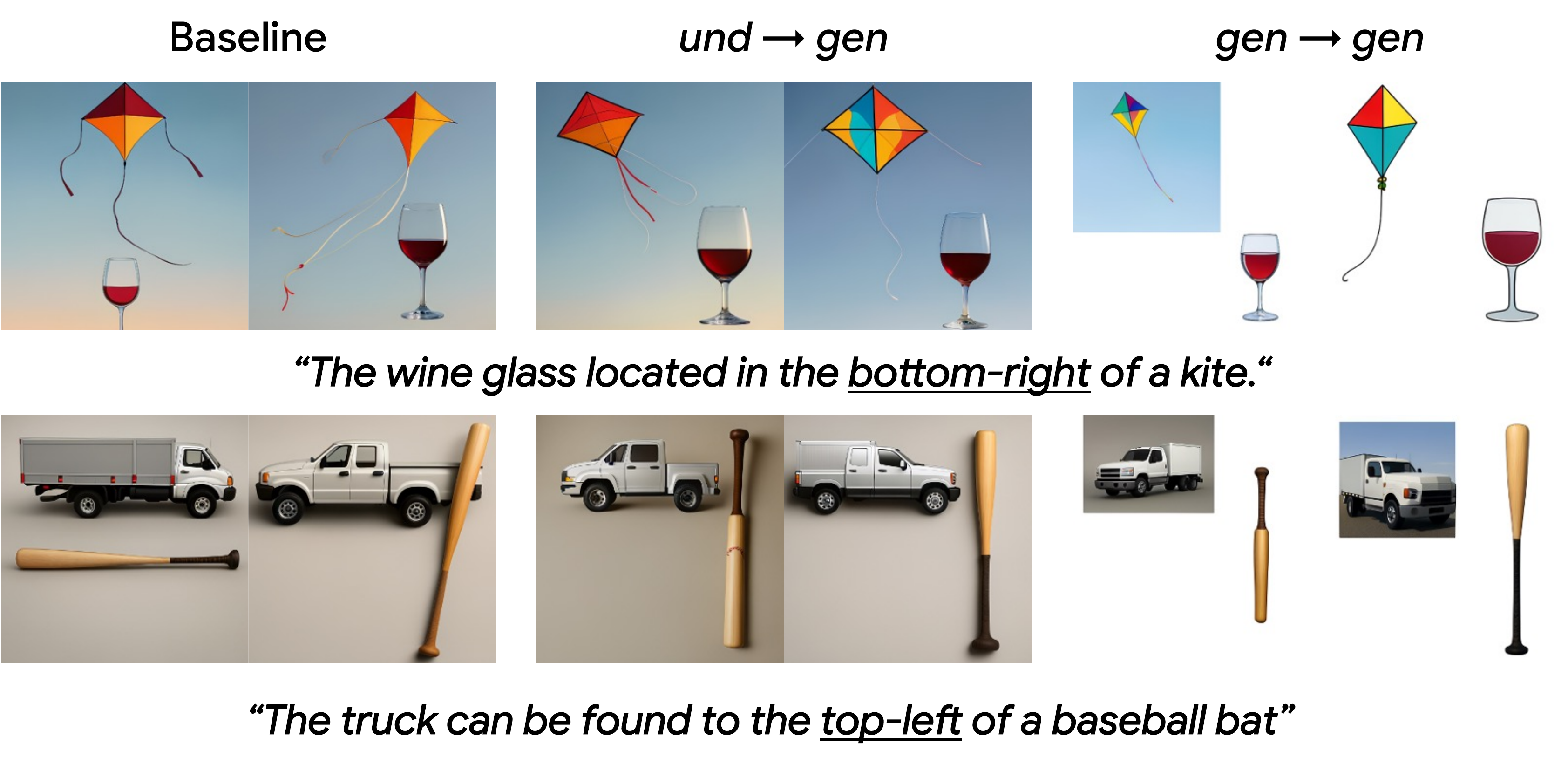} 
    \vspace{-20pt}
    \caption{
\textbf{Qualitative comparison between \undgen\ and \gengen\ on spatial relation tasks.} While generation-only training (\gengen) significantly degrades image quality reflecting traits of synthetic training data, \undgen\ preserves generation quality while achieving improved spatial relation accuracy.}
    \label{fig:spatial_pos}
\end{figure*}

\begin{table}[t]
    \centering
    \setlength{\tabcolsep}{6pt}
    \renewcommand{\arraystretch}{0.85}
    \begin{tabular}{c|cccc}
        \toprule
        \textbf{Train $\rightarrow$ Test}
        & \textbf{Acc.}~$\uparrow$
        & \textbf{IS}~\cite{salimans2016isscore}~$\uparrow$
        & \textbf{FID}~\cite{heusel2017fid}~$\downarrow$
        \\
        \midrule
        Baseline
        & 67.0 & 16.86 & - \\
        \undgen
        & 74.0 \good{+7.0}  & \textbf{17.50} &  \textbf{30.95}  \\
        \gengen
        & \textbf{80.0} \good{+13.0} &  17.41  &  32.28 \\
        \bottomrule
    \end{tabular}
    \vspace{6pt}

    \caption{\textbf{Quantitative comparison of understanding-to-generation transfer (\undgen) versus direct generation training (\gengen) on spatial relation.}
    Both \undgen\ and \gengen\ improve spatial accuracy over the baseline, with \undgen\ attaining a substantial gain (+7.0\%) from transferred understanding supervision alone, approaching direct generation training (+13.0\%). \undgen\ also stays closer to the baseline image distribution (lower FID)}

    \vspace{-20pt}
    \label{tab:gen_quality_spatial}
\end{table}

For evaluation, we use the same T2I prompt format as the training data and follow GenEval~\cite{ghosh2023geneval} protocol. We apply a detection model~\cite{cheng2022masked} to localize bounding boxes of the two objects and compute the center points of the detected boxes and verify whether their relative position matches the specified direction. More details of the dataset construction and evaluation pipeline can be found in the Appendix.

\subsubsection{Results.} 

Table~\ref{tab:gen_quality_spatial} confirms that understanding-to-generation transferability also holds for spatial relation. The model trained through understanding (\undgen) achieves higher spatial accuracy than the baseline, showing that relational knowledge transfers from understanding to generation. Training directly on generation data (\gengen) yields slightly higher accuracy but at the cost of a higher FID. Qualitative results in Figure~\ref{fig:spatial_pos} illustrate this trade-off: \gengen\ drifts toward the training distribution, producing images that resemble cropped objects pasted onto a canvas rather than natural scenes, whereas \undgen\ improves spatial correctness while preserving realistic appearance.

\subsection{Text recognition and generation}
\label{subsec:ocr}

\begin{wrapfigure}[13]{r}{0.33\linewidth}
    \centering
    \vspace{-20pt}
    \includegraphics[width=\linewidth]{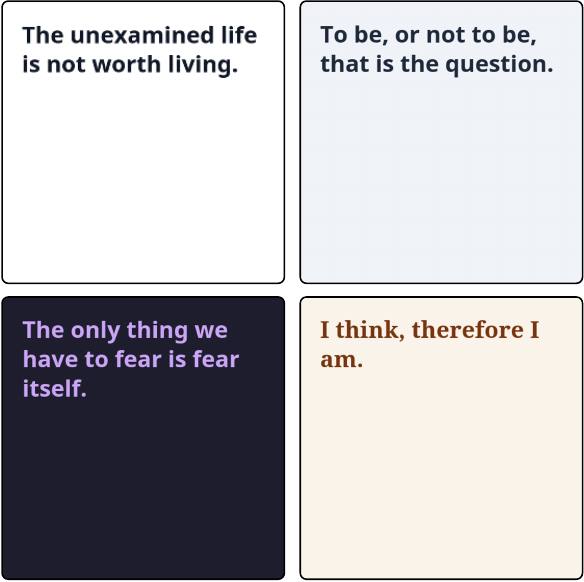}
    \vspace{-15pt}
    \caption{\textbf{Synthetic dataset for text recognition and generation.} }
    \label{fig:ocr_dataset}
    \end{wrapfigure}

We next examine text-contained image generation, where the model must render legible characters with accurate strokes, spacing, and glyph structure. Unlike counting or spatial relation, this requires the model to capture highly localized and fine-grained visual details. We investigate whether such fine-grained capabilities can be introduced through understanding and transferred into generation.

\begin{figure*}[t]
    \centering
    \includegraphics[width=\linewidth]{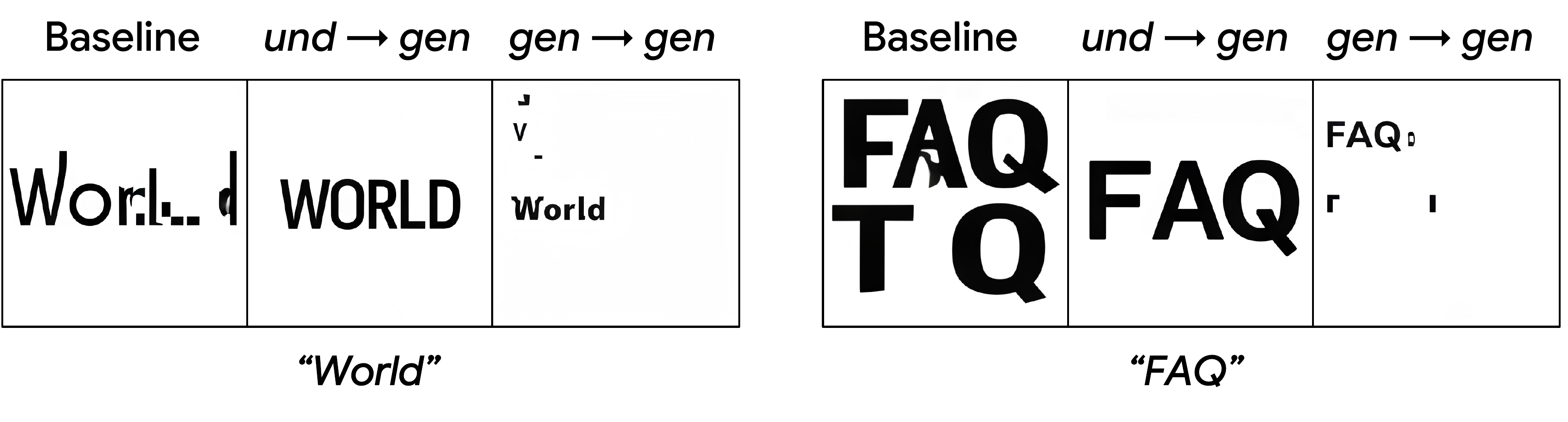} 
    \vspace{-20pt}
    \caption{\textbf{Qualitative comparison between \undgen\ and \gengen\ on text generation task.}  $und \rightarrow gen$ generates clear and accurate text, preserving rendering fidelity.}
    \label{fig:ocr_results}
    \vspace{-5pt}
\end{figure*}












\begin{table*}[!t]
\centering
\resizebox{\textwidth}{!}{%
    \begin{tabular}{c|cccccc}
    \toprule
    \textbf{Train $\rightarrow$ Test} 
    & \textbf{WER}~$\downarrow$
    & \textbf{CER}~$\downarrow$
    & \textbf{Edit Dist.}~$\downarrow$
    & \textbf{BLEU}~$\uparrow$
    & \textbf{METEOR}~$\uparrow$
    & \textbf{F1}~$\uparrow$
    \\
    \midrule
    Baseline & 0.654 & 0.362 & 36.812 & 0.277 & 0.403 & 0.446 \\
    \midrule
    \undgen
    & 0.641 \good{-0.013} & 0.351 \good{-0.011} & 35.636 \good{-1.176} & 0.274 \bad{-0.003} & 0.410 \good{+0.007} & 0.452 \good{+0.006} \\
    \gengen
    & 0.641 \good{-0.013} & 0.367 \bad{+0.005} & 37.312 \bad{+0.500} & 0.354 \good{+0.077} & 0.533 \good{+0.130} & 0.525 \good{+0.079} \\
    \bottomrule
    \end{tabular}
}
\vspace{5pt}
\caption{\textbf{Quantitative comparison of understanding-to-generation transfer (\undgen) versus direct generation training (\gengen) on text generation.} We apply an OCR model~\cite{glm_ocr} to the generated images for text extraction and compute WER, CER, edit distance, BLEU, METEOR, F1 between the extracted text and the original prompt. \undgen\ transfer yields improvements in most metrics.}
\vspace{-10pt}
\label{tab:ocr_transfer}
\end{table*}

\subsubsection{Experimental setting.} We find that existing OCR datasets tend to focus on either too small, sign-level text in natural scenes~\cite{singh2021textocr, veit2016coco, UberText} or dense, full-page documents~\cite{nemotron_dataset}. Both may be too hard for models based on discrete image tokenizers, as quantizing such images into discrete tokens discards much of the fine-grained visual details. Thus, we construct a synthetic dataset of 200K image-text pairs by rendering Markdown documents into images~(Figure~\ref{fig:ocr_dataset}). Our rendered dataset provides diverse text content at a moderate difficulty level. For evaluation, we apply an OCR model~\cite{glm_ocr} to read text from the generated image and compute five metrics against the original prompt: word error rate (WER) and character error rate (CER) measure exact correctness at the word and character level. Edit Distance captures the minimum number of operations needed to align the prediction with the ground truth. METEOR~\cite{banerjee2005meteor} and F1 assess partial matches by accounting for token overlap and ordering. Details of the dataset construction and evaluation pipeline can be found in the Appendix.

\subsubsection{Results.} Table~\ref{tab:ocr_transfer} summarizes the results. The model trained through understanding~(\undgen) shows improvements over the baseline on most metrics, including WER, CER, Edit Distance, METEOR and F1, confirming that understanding-to-generation transfer also exists for text generation. The qualitative examples in Figure~\ref{fig:ocr_results} further verify this: the understanding-trained model produces more legible text compared to the baseline. 
While direct training with generation objective~(\gengen) yields larger gains in text accuracy, this comes at the cost of generation quality. 
These results together indicate that the transfer-based approach remains a safer alternative that avoids degrading generative quality when boosting specific visual capabilities. 

\section{Discussion}
\label{sec:discussion}
While our experiments demonstrate that understanding-to-generation transfer can effectively improve generative capabilities, several natural questions remain. In this section, we further analyze the properties and implications of transferability in UMMs. We first investigate whether the reverse direction, \ie, generation-to-understanding transfer, can also provide meaningful improvements. We then examine whether jointly training understanding and generation tasks could serve as an alternative to transfer-based training. Finally, we analyze whether our understanding training negatively affects general multimodal understanding ability.

\subsection{Can generation-to-understanding transfer also be useful?}
\label{sec:discuss_ocr}

While the previous section demonstrated that \undgen\ transfer can be exploited to improve specific generative capabilities, we observed that the strength of transferability is not uniform across capabilities. To quantify the strength, we assume direct (\eg, generation) training on the same dataset as the upper bound for transfer (\eg, \undgen). We then measure transfer strength as the ratio of the improvement from transfer to that from direct training, with the results presented in Table~\ref{tab:transfer_strength}. We found that for text generation, the transfer strength of \undgen\ is overall lower than that of counting and spatial relation. This raises the question of what factors determine how well a capability transfers.

\begin{table*}[t]
\centering
\resizebox{\textwidth}{!}{%
    \begin{tabular}{c|cc|c|cccccc}
    \toprule
    \multirow{2}{*}{\textbf{Train $\rightarrow$ Test}}
    & \multicolumn{2}{c|}{\textbf{Counting}}
    & \multicolumn{1}{c|}{\textbf{Spatial Rel.}}
    & \multicolumn{6}{c}{\textbf{Text Recognition/Generation}} \\
    \cmidrule(lr){2-10}

    & \textbf{Acc. (\%)}~$\uparrow$ & \textbf{MAD}~$\downarrow$
    & \textbf{Acc. (\%)}~$\uparrow$
    & \textbf{WER}~$\downarrow$ & \textbf{CER}~$\downarrow$ & \textbf{Edit Dist.}~$\downarrow$
    & \textbf{BLEU}~$\uparrow$ & \textbf{METEOR}~$\uparrow$ & \textbf{F1}~$\uparrow$ \\
    \midrule
    Baseline
    & 22.0 & 2.48
    & 61.0
    & 0.129 & 0.073 & 7.464 & 0.835 & 0.899 & 0.885 \\
    \midrule
    \genund
    & 30.0 \good{+8.0} & 1.88 \good{-0.60}
    & 67.0 \good{+6.0}
    & 0.118 \good{-0.011} & 0.061 \good{-0.012} & 6.232 \good{-1.232} & 0.845 \good{+0.010} & 0.907 \good{+0.008} & 0.892 \good{+0.007} \\
    \undund
    & 57.0 \good{+35.0} & 0.89 \good{-1.59}
    & 89.0 \good{+28.0}
    & 0.091 \good{-0.038} & 0.042 \good{-0.031} & 4.276 \good{-3.188} & 0.875 \good{+0.040} & 0.928 \good{+0.029} & 0.912 \good{+0.027} \\
    \bottomrule
    \end{tabular}
} 
\vspace{5pt}
\caption{\textbf{Generation-to-understanding transfer (\genund) versus direct understanding training (\undund) on counting, spatial relation, and text recognition for Lumina-DiMOO.} Across all three capabilities, \textit{gen}~$\rightarrow$~\textit{und} transfer consistently improves over the baseline.}
\label{tab:ocr_transfer_genund}
\vspace{-10pt}
\end{table*}

\begin{table}[!t]
\centering
\setlength{\tabcolsep}{3pt}
\resizebox{0.85\width}{!}{%
\begin{tabular}{c|cc|c|ccc}

\toprule
\multirow{2}{*}{\textbf{Transfer Direction}}
& \multicolumn{2}{c|}{\textbf{Counting}}
& \multicolumn{1}{c|}{\textbf{Spatial Rel.}}
& \multicolumn{3}{c}{\textbf{Text Recog./Gen.}} \\

& Acc. & MAD & Acc. & WER & METEOR & F1 \\
\midrule


\undgen
& \textbf{100\%} & \textbf{140\%} & \textbf{53.8\%} & \textbf{100\%} & 5.4\% & 7.6\% \\

\midrule

\genund
& 22.9\% & 37.7\% & 21.4\% & 28.9\% & \textbf{27.6\%} & \textbf{25.9\%} \\

\bottomrule

\end{tabular}
}

\vspace{5pt}

\caption{
    \textbf{Transfer strength comparison across tasks and directions.} Assuming direct training as an upper bound that transfer can achieve, we quantify transfer strength as the ratio of the improvement obtained by transfer to that obtained by direct training. For counting and spatial-relation, \undgen\ transfer is stronger than \genund, whereas for text recognition/generation the opposite holds, indicating that the dominant transfer direction may be capability-dependent.
}

\vspace{-15pt}
\label{tab:transfer_strength}
\end{table}

We hypothesize that the strength of \undgen\ transfer depends on the nature of the visual knowledge each capability requires. Counting and spatial relation rely on relatively high-level, structural concepts, such as numerical quantity and relative object layout. Consequently, transferring from understanding to generation for these tasks does not strictly require pixel-level precision. The text capability, by contrast, intrinsically demands highly fine-grained visual details, such as stroke geometry, character spacing, and glyph composition. Such fine-grained precision is difficult to acquire purely through the understanding objective~\cite{zhang2023visual,zhang2024exploring,khayatkhoei2025mllms,jung2026visual,ghosh2026understanding}, which often relies on partial or contextual cues to perform recognition without fully encoding the exact visual details needed for rendering. This gap between the coarse-grained representations learned through the understanding objective~\cite{marsili2026same, wu2025harmonizing} and the fine-grained precision inherently required by the text capability explains the weaker \undgen\ transfer.

This reasoning leads to a natural follow-up question: if text generation requires pixel-level precision, and text recognition fundamentally demands the same level of detailed visual discrimination, does the \textit{reverse transfer direction (\genund) work more effectively} for such capabilities? Since the generation objective inherently provides more detailed supervision than the understanding objective, we expect that learning to generate text forces the model to capture the fine-grained representations highly beneficial for text recognition. 


To investigate this, we extend our evaluation to the remaining capabilities---text and spatial relation---by reporting the \genund\ transfer performance on Lumina-DiMOO alongside direct training (\undund) for comparison (Table~\ref{tab:ocr_transfer_genund}), and calculate their respective transfer strengths (Table~\ref{tab:transfer_strength}). Consistent with our hypothesis, Table~\ref{tab:transfer_strength} demonstrates that while \undgen\ transfer remains stronger for counting and spatial relation, the reverse direction (\genund) more robustly improves performance for text capabilities compared to \undgen. These findings confirm that the dominant direction and strength of transferability are capability-dependent and governed by the granularity of supervision required.

\subsection{Does joint training lead to stronger transferability?}
\label{sec:joint_training}

\begin{figure}[t]
    \centering
    \includegraphics[width=\linewidth]{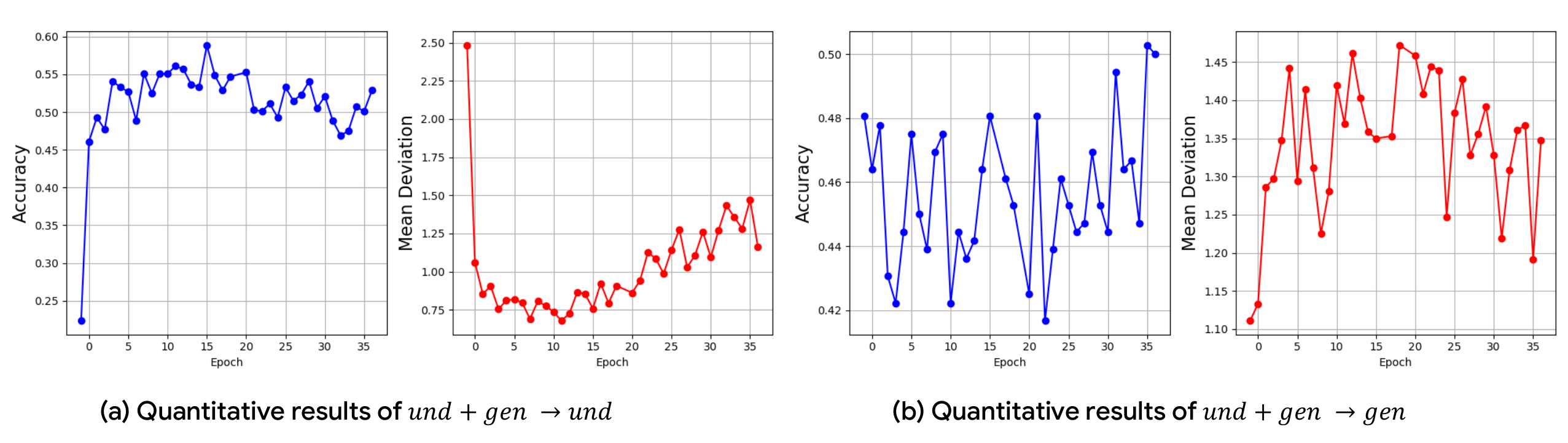}
    \vspace{-15pt}
    \caption{\textbf{Trade-off in joint \textit{und}+\textit{gen}\ training.} (a) Counting understanding accuracy and mean absolute deviation (MAD) over training epochs. (b) Counting generation accuracy and MAD over training epochs. Blue lines denote accuracy and red lines denote MAD.}
    \label{fig:joint_training}
    \vspace{-10pt}
\end{figure}

Throughout this work, we examined transferability in UMMs and showed that training a capability through one task can improve it in the other task. This naturally raises the question of whether training both tasks simultaneously leads to improved performance for both.

To investigate this, we trained Lumina-DiMOO on understanding and generation jointly (\textit{und} + \textit{gen}) for counting. Results shown in Figure~\ref{fig:joint_training} indicate that the two objectives do not improve simultaneously during training. Instead, their performance exhibits an oscillatory pattern. Understanding accuracy increases in the first half of training while generation accuracy declines, and the trend reverses in the latter half. This behavior suggests that the presence of transferability does not necessarily translate into consistent gains when the two are optimized simultaneously.

One possible explanation is that understanding and generation require different alignment patterns within the network, which may lead to conflicting optimization signals during joint training, as suggested by~\cite{li2025unifork}. At the same time, our experimental setting may limit how clearly synergy can be observed. We conduct our experiments as post-training on a pretrained model whose capabilities are already largely established, which may lead to different training dynamics from those that arise during large-scale pre-training. In addition, the joint training uses the same concept-focused dataset constructed for the individual experiments in Section~\ref{sec:umms}, where each image-text pair is tightly aligned around \textit{counting}. Such paired data distributions are uncommon in typical multimodal pre-training corpora and may introduce an unfamiliar training configuration when the two objectives are optimized together. Distinguishing between these possibilities is difficult within the scope of post-training experiments and therefore remains an open question.

\subsection{Does understanding training harm general multimodal ability?}



\begin{table}[t]
    \centering
    \begin{tabular}{c|cccc}
        \toprule
       \rowcolor{white}
        \textbf{Train (\textit{und})} 
        & \textbf{POPE}~\cite{li2023pope}
        & \textbf{MMBench}~\cite{liu2024mmbench}
        & \textbf{MMMU}~\cite{yue2024mmmu}
        & \textbf{MME-P}~\cite{fu2023mme}
        \\
        \midrule
        Baseline & 86.7 & 90.8 & 57.7 & 1534\\
        Counting & 86.4 & 90.9 & 62.0 & 1492\\
        Spatial Relation & 87.3 & 90.8 & 61.3 & 1554\\
        Text recognition & 86.7 & 90.5 & 61.3 & 1538 \\
        \bottomrule
    \end{tabular}
    \vspace{6pt}
    \caption{\textbf{Performance on general multimodal benchmarks after understanding-based training.} Despite introducing task-specific capabilities through understanding training for \textit{und}~$\rightarrow$~\textit{gen} transfer, the model's general multimodal understanding remains largely preserved across all benchmarks.}
    \label{tab:general_bench}
    \vspace{-20pt}
\end{table}




A potential concern of capability injection through understanding training is that it may negatively affect the model’s overall multimodal ability. Since our approach introduces additional supervision targeting specific capabilities, one might expect such training to overfit to the target task or degrade performance on general multimodal benchmarks.

To examine this, we evaluate the models on several widely used multimodal understanding benchmarks, including POPE~\cite{li2023pope}, MMBench~\cite{liu2024mmbench}, MMMU~\cite{yue2024mmmu}, and MME~\cite{fu2023mme}. The results are summarized in Table~\ref{tab:general_bench}. Across all benchmarks, we observe that understanding-based training does not degrade general multimodal performance. The scores on POPE and MMBench remain nearly unchanged compared to the baseline model, while performance on MMMU even improves slightly after capability-focused training. These results suggest that injecting capabilities through understanding tasks introduces minimal changes to the model parameters while preserving the model’s general multimodal knowledge.

\section{Conclusion}


In this work, we studied cross-task transferability between understanding and generation in UMMs. Through controlled experiments, we empirically showed that capabilities transfer between the two tasks and that the strength of this transfer depends on architectural design. We further exploited this as a practical strategy: introducing a target capability through understanding, rather than fine-tuning generation directly, improves capability-specific performance while preserving overall generative quality. We hope our findings provide new insights into the interaction between understanding and generation and open promising directions for post-training in unified multimodal models.

\section*{Acknowledgements}
This research was supported by Institute of Information \& Communications Technology Planning \& Evaluation (IITP) grant funded by the Korea government (MSIT) (RS-2019-II190075, RS-2024-00509279, RS-2025-II212068, RS-2023-00227592, RS-2025-02214479, RS-2024-00457882, RS-2025-25441838, RS-2025-02217259, RS-2026-25519202) and the Culture, Sports, and Tourism R\&D Program through the Korea Creative Content Agency grant funded by the Ministry of Culture, Sports and Tourism (RS-2024-00345025, RS-2024-00333068), and National Research Foundation of Korea (RS-2024-00346597).

%

%
\bibliographystyle{splncs04}
\bibliography{main}

\clearpage
\appendix
\setcounter{page}{1}
\setcounter{linenumber}{1}
\etocdepthtag.toc{appendix} 
\section*{\Large Appendix}


\makeatletter
\newcommand{\appendix@noopaddcontentsline}[3]{}
\let\appendix@origsection\section
\renewcommand{\section}{\@ifstar\appendix@origsection\appendix@section}
\newcommand{\appendix@section}[1]{%
  \let\appendix@savedaddcontentsline\addcontentsline
  \let\addcontentsline\appendix@noopaddcontentsline
  \appendix@origsection{#1}%
  \let\addcontentsline\appendix@savedaddcontentsline
  \phantomsection
  \addcontentsline{toc}{section}{\protect\numberline{\thesection}#1}%
}
\let\appendix@origsubsection\subsection
\renewcommand{\subsection}{\@ifstar\appendix@origsubsection\appendix@subsection}
\newcommand{\appendix@subsection}[1]{%
  \let\appendix@savedaddcontentsline\addcontentsline
  \let\addcontentsline\appendix@noopaddcontentsline
  \appendix@origsubsection{#1}%
  \let\addcontentsline\appendix@savedaddcontentsline
  \phantomsection
  \addcontentsline{toc}{subsection}{\protect\numberline{\thesubsection}#1}%
}
\makeatother

\etocsetlevel{author}{6}
\etocsetlevel{title}{6}
\etocsettagdepth{main}{none}
\etocsettagdepth{appendix}{subsection}
\makeatletter
{%
\let\section\appendix@origsection
\let\subsection\appendix@origsubsection
\tableofcontents
}
\makeatother

\etocdepthtag.toc{appendix} 

\clearpage
 
\section{Training Details}
\label{sec:appendix_experimental_details}

In the experiments of Section~\ref{sec:umms}, we evaluate four representative UMMs spanning different architectural families (Figure~\ref{fig:arch_categorize}): BAGEL~\cite{deng2025bagel} (7B), Janus-Pro~\cite{chen2025januspro} (7B), Lumina-DiMOO~\cite{xin2025lumina} (8B), and BLIP3-o~\cite{chen2505blip3} (8B). All models fall within the 7B–8B parameter range to enable a fair comparison across architectures. To investigate whether transferability emerges in the shared LLM backbone, we perform LoRA-based supervised fine-tuning on top of publicly available pretrained weights. We apply LoRA with rank 128 to the query, key, and value projection layers as well as the MLP layers of the transformer blocks. All other components—including the image encoder, projector/aligner, language modeling head, and diffusion head—are kept frozen. We use 8-bit AdamW with bfloat16 training across all runs. We train for 15 epochs with early stopping to get the best checkpoint and use one capability-specific dataset per training run without additional data balancing. Unless otherwise specified, we follow the official codebases and recommended training configurations of each model. The model-specific settings are described below.

\noindent\textbf{BAGEL~\cite{deng2025bagel}.}
We use a learning rate of $1\times10^{-4}$ and retain the model’s original dynamic image resizing strategy.

\noindent\textbf{Janus-Pro~\cite{chen2025januspro}.}
We use a learning rate of $4\times10^{-5}$. For both understanding and generation, we preserve the aspect ratio, resize the longer side to 384 pixels, and pad the shorter side, resulting in a $384\times384$ input.

\noindent\textbf{Lumina-DiMOO~\cite{xin2025lumina}.}
We use a learning rate of $3\times10^{-6}$. For both understanding and generation, we preserve the aspect ratio, resize the longer side to 512 pixels, and pad the shorter side, resulting in a $512\times512$ input. We use the same training configuration for the experiments in Sections~\ref{sec:tasks} and~\ref{sec:discuss_ocr}.

\noindent\textbf{BLIP3-o~\cite{chen2505blip3}.}
We use a learning rate of $2\times10^{-5}$. Since BLIP3-o employs separate transformer branches for understanding and generation, we fine-tune only the branch corresponding to each task. We retain the original image processing strategy of Qwen2.5-VL-7B-Instruct~\cite{bai2025qwen2}, which serves as the LLM backbone of BLIP3-o.

\clearpage
\section{Task Details}
\label{sec:appendix_dataset_construction}
\subsection{Counting}
\subsubsection{Dataset details.} Our \textit{visual instruction tuning dataset} for counting consists of 65.6K images and approximately 200K question-answer pairs. We derive the VQA pairs from the training sets of PixMo-Count~\cite{deitke2025molmo} and PixMo-Points~\cite{deitke2025molmo}.
For counting understanding, we directly adopt the original question-answer pairs from PixMo-Count, as it was originally designed for counting tasks. To scale up the training data, we additionally incorporate samples from PixMo-Points. Since PixMo-Points is a point prediction dataset rather than a counting dataset, we reformulate it for counting by treating the number of points to be predicted as the object count. Note that PixMo-Points originally contains annotations of more than 1K points per image. We find that such extreme cases introduce excessive difficulty during training. To maintain a reasonable difficulty level, we filter out samples whose object count exceeds 20. For the \textit{counting-aware generation dataset}, we construct the training set from the same 65.6K images used in the understanding split to ensure a controlled experimental setting. Specifically, we leverage the object count annotations from the understanding training set and use Qwen3-VL~\cite{bai2025qwen3} to generate count-aware generation descriptions conditioned on these annotations.
Table~\ref{tab:dataset_counting} presents representative samples from both the counting understanding and generation training sets.
\begin{table*}[t]
\centering
\setlength{\tabcolsep}{6pt}
\resizebox{\textwidth}{!}{%
\begin{tabular}{cccc}
\toprule
{\textbf{Image}} & {\textbf{PixMo-Points/Counts}} & \textbf{Understanding QA} &{\textbf{Description}} \\
\midrule
\addlinespace[4pt]
\raisebox{-0.5\height}{\includegraphics[width=5cm]{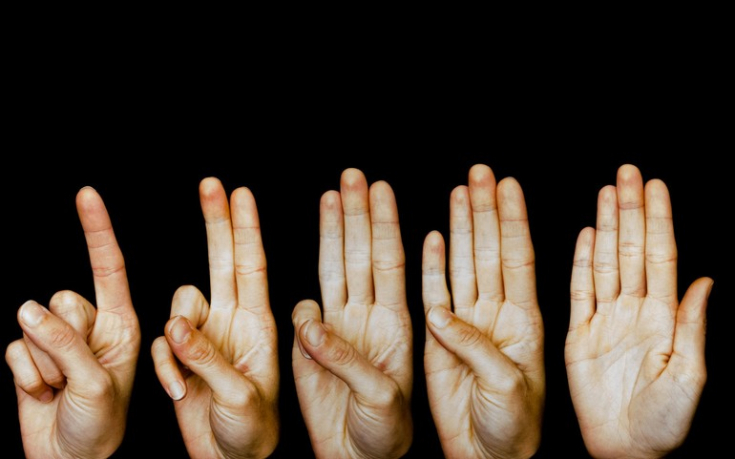}} &
\parbox[c]{4cm}{\raggedright\scriptsize <points x1="9.388" y1="75.817" x2="26.667" y2="78.214" x3="47.483" y3="80.392" x4="65.578" y4="77.124" x5="93.469" y5="73.638" alt="thumbs">, <points x1="89.932" y1="57.081" x2="70.612" y2="56.209" x3="51.429" y3="56.427" x4="34.558" y4="60.784" x5="14.014" y5="59.477" alt="index fingers"> \ldots
} &
\parbox[c]{4cm}{\raggedright\scriptsize Q: How many thumbs are there in the image? Response Example : There are <number> thumbs in the image. \\[8pt]
A: There are ** 5 ** thumbs in the image.} &
\parbox[c]{6cm}{\raggedright\scriptsize 5 thumbs, 5 index fingers, 5 middle fingers, 5 pinky fingers, 5 number of right thumbs, 5 right pointer fingers are arranged in a horizontal row against a solid black background. Each hand displays exactly five fingers, including the thumb, with the right thumb and right pointer finger clearly visible on each. The skin tones vary slightly, and the lighting highlights the natural texture and creases of the palms and fingers. A total of five hands are shown, each positioned to emphasize the count of fingers and thumbs, with no additional objects present.} \\
\addlinespace[10pt]
\raisebox{-0.5\height}{\includegraphics[width=5cm]{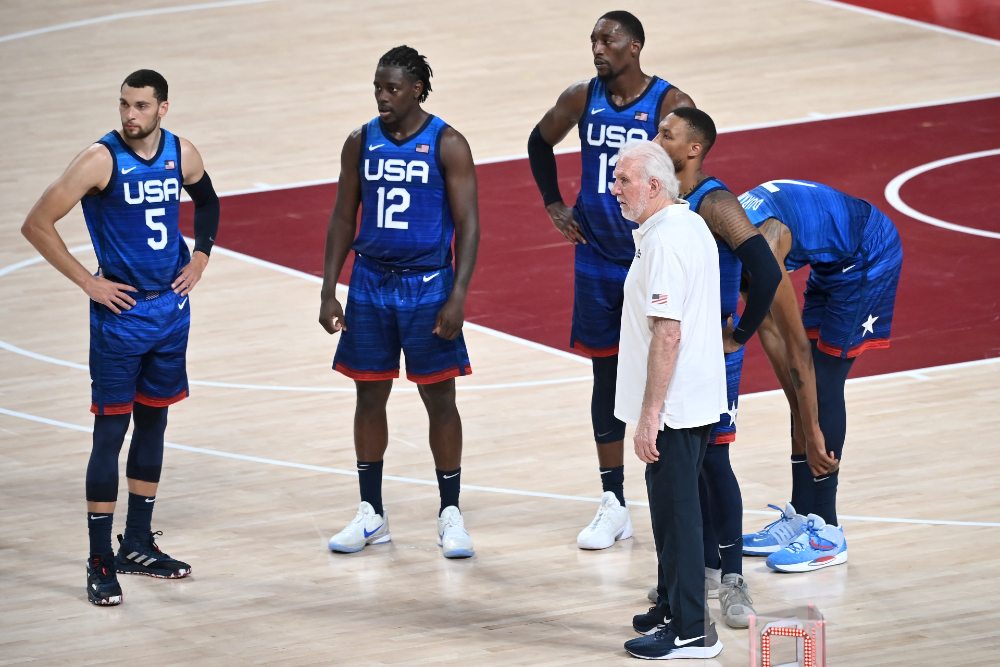}} &
\parbox[c]{4cm}{\raggedright\scriptsize <points x1="81.15" y1="35.082" x2="67.25" y2="46.327" x3="68.85" y3="21.589" x4="62.65" y4="11.094" x5="42.75" y5="34.933" x6="16.35" y6="38.081" alt="people">, <points x1="68.35" y1="20.69" x2="63.95" y2="25.937" x3="61.45" y3="6.147" x4="39.15" y4="11.394" x5="14.75" y5="14.543" alt="heads"> \ldots} &
\parbox[c]{4cm}{\raggedright\scriptsize Q: How many people are there in the image? Response Example : There are <number> people in the image. \\[8pt]
A: There are ** 6 ** people in the image.} &
\parbox[c]{6cm}{\raggedright\scriptsize Six people, five heads, six people on court, eleven shoes, ten arms, nine socks, six humans, nine letters, seven hands, five numbers, eight eyes, six white lines, nine elbows, five shorts. The scene shows exactly six people on court, each wearing blue uniforms with white lettering and numbers, standing on a wooden floor marked with six white lines. A total of eleven shoes are visible, paired with nine socks, and the players’ arms and elbows are clearly defined. The uniforms feature five numbers and nine letters, with eight eyes and seven hands visible. Five shorts are worn by the players, and the composition is lit evenly under bright overhead lighting.} \\
\addlinespace[10pt]
\raisebox{-0.5\height}{\includegraphics[width=5cm]{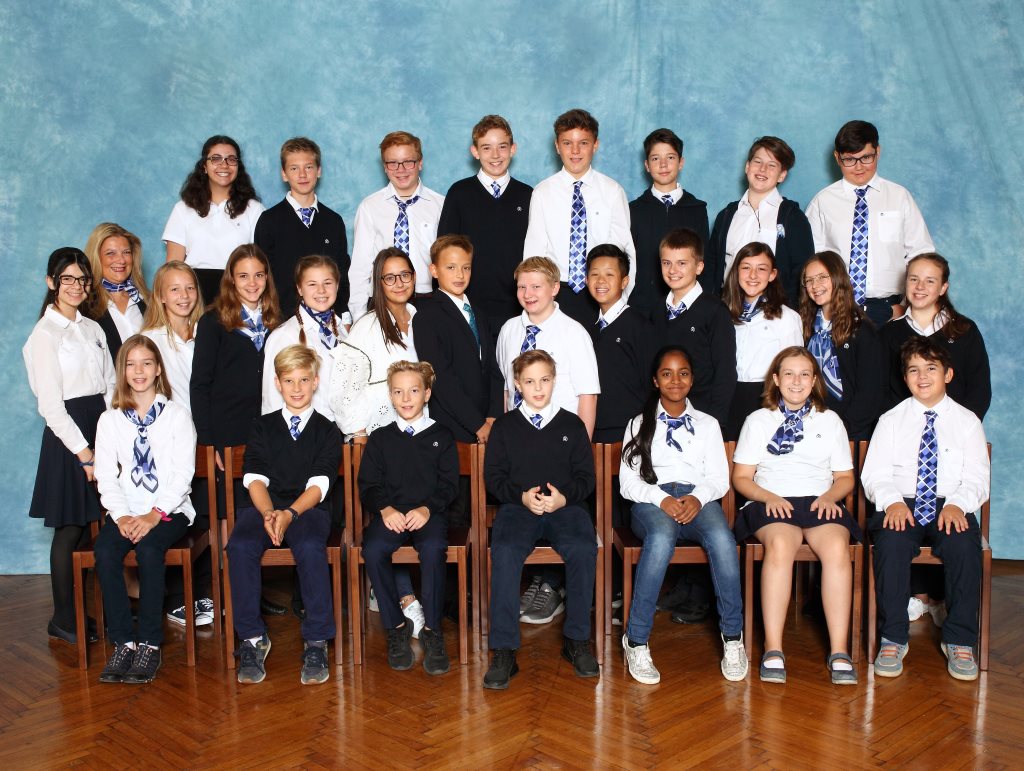}} &
\parbox[c]{4cm }{\raggedright\scriptsize <points x1="15.918" y1="63.424" x2="28.613" y2="65.37" x3="42.09" y3="66.796" x4="51.367" y4="64.591" x5="63.477" y5="66.537" x6="75.977" y6="65.24" x7="87.695" y7="65.37" alt="people sitting">, <points x1="22.363" y1="23.476" x2="89.746" y2="39.559" \ldots  alt="female"> \ldots} &
\parbox[c]{4cm }{\raggedright\scriptsize Q: How many people sitting are there in the image? Response Example : There are <number> people sitting in the image. \\[8pt]
A: There are ** 7 ** people sitting in the image.} &
\parbox[c]{6cm}{\raggedright\scriptsize 7 people sitting, 13 female, 15 males, 6 people with glasses, 14 ties, 8 scarves, 5 people wearing glasses, 7 wooden chairs, 13 Females, 7 Sitting people. The group is arranged in three rows, with exactly 7 people sitting on 7 wooden chairs. A total of 13 females and 15 males are visible, wearing white shirts and blue patterned ties or scarves. Exactly 6 people have glasses, and 5 of them are wearing glasses, with 14 ties and 8 scarves distributed among them. The background is a textured blue wall, and the polished wooden floor reflects the studio lighting.} \\
\addlinespace[10pt]
\raisebox{-0.5\height}{\includegraphics[width=5cm] {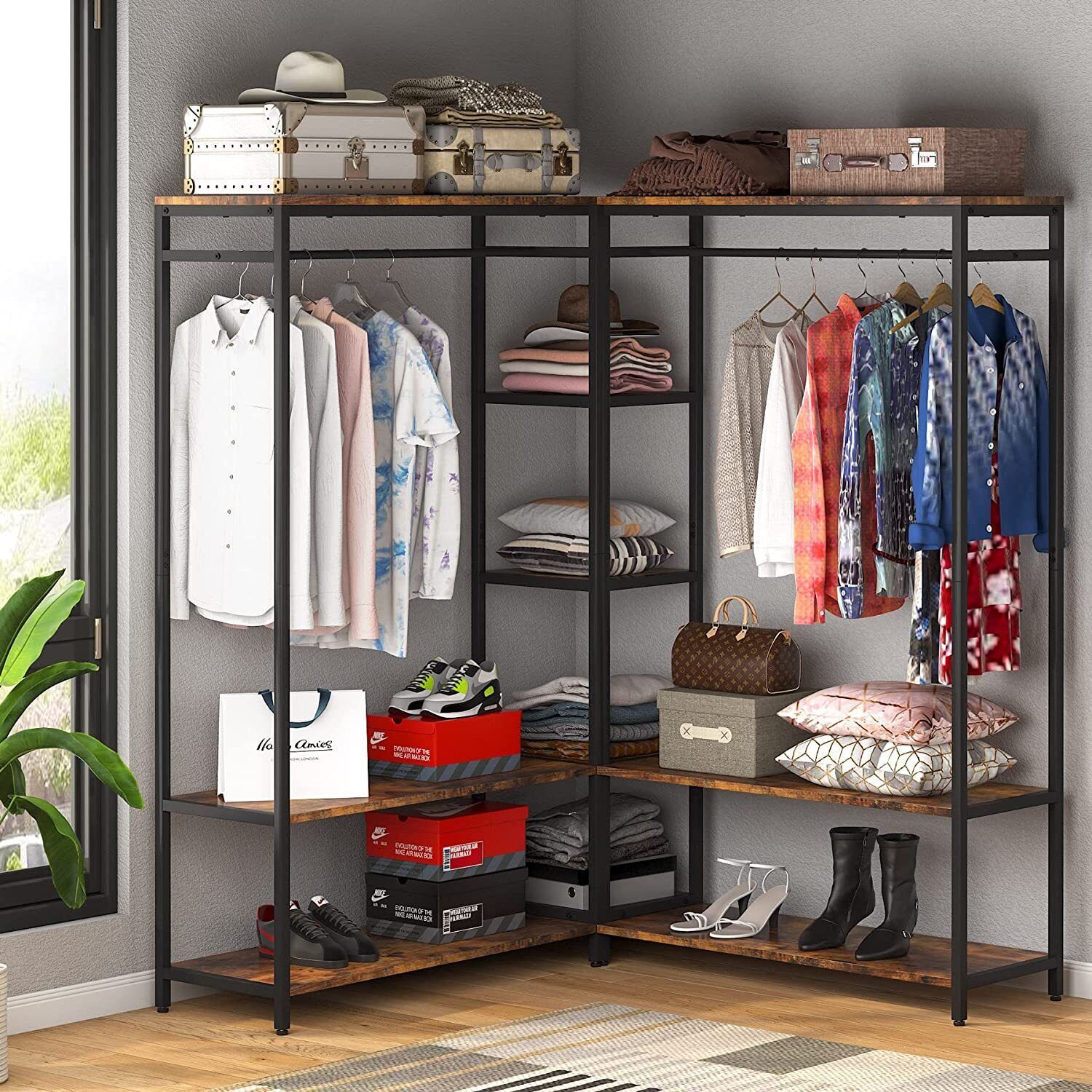}} &
\parbox[c]{4cm }{\raggedright\scriptsize <points x1="57.325" y1="47.714" x2="58.008" y2="50.741" x3="85.059" y3="65.194" x4="82.813" y4="70.468" alt="pillows">, <points x1="42.188" y1="67.782" x2="42.286" y2="76.962" x3="42.872" y3="83.602" x4="49.415" y4="81.356" x5="65.919" y5="67.489" alt="boxes"> \ldots} &
\parbox[c]{4cm }{\raggedright\scriptsize Q: How many pillows are there in the image? Response Example : There are <number> pillows in the image. \\[8pt]
A: There are ** 4 ** pillows in the image.} &
\parbox[c]{6cm}{\raggedright\scriptsize 4 pillows, 5 boxes, 8 shoes, 11 hangers, 6 leaves, 13 folded items, 15 folded textile items, 10 shirts hanging up, 11 hanger body, 11 metal hooks, 6 leafs, 9 grey buttons on white shirt on end, 10 clothes on clothes hangers. The metal-framed shelving unit holds exactly 10 shirts hanging up and a total of 15 folded textile items. Four pillows rest on lower shelves, and exactly 8 shoes are arranged neatly. A total of 5 boxes, including one red and one black, sit on the bottom shelf. Six leaves are visible near the window, and the 11 hangers support 10 clothes on clothes hangers.} \\
\addlinespace[4pt]
\bottomrule
\end{tabular}
}
\caption{Examples of counting dataset.}
\label{tab:dataset_counting}
\end{table*}

\subsubsection{Understanding evaluation.}
We evaluate counting understanding using 540 QA pairs constructed from 
the test split of PixMo-Count. Each prompt includes a \texttt{Response Example} 
field that instructs the model to place the predicted count inside markdown 
markers \texttt{** **}. Each QA pair follows the template:
\begin{quote}
\textbf{Question:} \texttt{"How many \{OBJECT\}s are there in the image? Response Example: There are **<COUNT>** of \{OBJECT\}s in the image."} \\
\textbf{Answer:} \texttt{"There are **\{COUNT\}** of \{OBJECT\}s in the image."}
\end{quote}
\noindent During evaluation, we parse the substring enclosed by the markers 
and compare it with the ground-truth count. If parsing fails, we fall back 
to checking whether the ground-truth count appears anywhere in the model 
output, either as an Arabic numeral or as an English number word.

\subsubsection{Generation evaluation.} 
Following the prompt generation protocol of GenEval~\cite{ghosh2023geneval}, 
we construct a total of 90 prompts by combining 10 object classes 
and 9 count words.
Each prompt follows the template:

\begin{quote}
\texttt{``a photo of \{COUNT\} \{OBJECT\}s''}
\end{quote}

\noindent where \texttt{OBJECT} $\in$ \{\texttt{person, dog, cat, car, 
cup, chair, book, bottle, apple, tie}\} and \texttt{COUNT} $\in$ 
\{\texttt{two, three, four, five, six, seven, eight, nine, ten}\}.

For each prompt, we generate four images with different random seeds 
and report the average accuracy across all generated samples.
To obtain object counts from generated images, we follow the same detection-based approach as GenEval~\cite{ghosh2023geneval}.

\subsection{Spatial Relation}
\subsubsection{Dataset details.}
Our \textit{visual instruction tuning dataset} for spatial relation understanding consists of 200K image-question-answer pairs. We construct composite images using object crops drawn from the ImageNet-1K~\cite{deng2009imagenet} training set. Specifically, for each sample, we randomly select two images from different ImageNet classes and paste them onto a $512 \times 512$ canvas with a random color background. We resize each selected image to a random size~(between 150 and 300 pixels per side, with an aspect ratio constrained near 1:1) and place them such that the two images do not overlap and exhibit a clearly unambiguous relative spatial relationship. 

We consider four relative positions—\texttt{top-left}, \texttt{top-right}, \texttt{bottom-left}, and \texttt{bottom-right}—defined by the difference between the centroids of the two images, and enforce an even distribution of 50K samples per position. For each sample, a question-answer pair is generated by randomly selecting one of 10 diverse QA templates that ask about the relative position of object $A$ with respect to object $B$. Each question is appended with a \texttt{Response format} hint that specifies the expected answer structure. For the \textit{spatial-relation-aware generation dataset}, we construct the training set from the same images used in the understanding split. We leverage the spatial relation annotations from the understanding training set and use Qwen3-VL~\cite{bai2025qwen3} to generate spatially-aware generation prompts conditioned on these annotations. Table~\ref{tab:spatial_dataset} presents representative samples from both the spatial relation understanding and generation training sets.

Each QA pair follows the template: 
\begin{quote}
\textbf{Question:} \texttt{"\{QUESTION\}\textbackslash nResponse format: \{RESPONSE\_FORMAT\}"} \\ 
\textbf{Answer:} \texttt{"\{ANSWER\}"} 
\end{quote} 

\noindent For example, given a template sampled at random: \begin{quote} \textbf{Question:} \texttt{"Where is the \{OBJECT\_A\} located relative \\ to the \{OBJECT\_B\}?} \\ \texttt{Response format: The \{OBJECT\_A\} is located in the \\ top-left/top-right/bottom-left/bottom-right of the \{OBJECT\_B\}."} \\ \textbf{Answer:} \texttt{"The \{OBJECT\_A\} is located in the \{POSITION\} of the \{OBJECT\_B\}."} \end{quote} 

\noindent where \texttt{POSITION} $\in$ \{\texttt{top-left, top-right, bottom-left, bottom-right}\}, and \texttt{OBJECT\_A}, \texttt{OBJECT\_B} are ImageNet class names.
\begin{table*}[t]
\centering
\label{tab:spatial_example}
\setlength{\tabcolsep}{6pt}
\resizebox{\textwidth}{!}{%
\begin{tabular}{cccc}
\toprule
{\textbf{Image}} & {\textbf{Prompt}} & {\textbf{Image}} & {\textbf{Prompt}} \\
\midrule
\addlinespace[4pt]
\raisebox{-0.5\height}{\includegraphics[width=5cm]{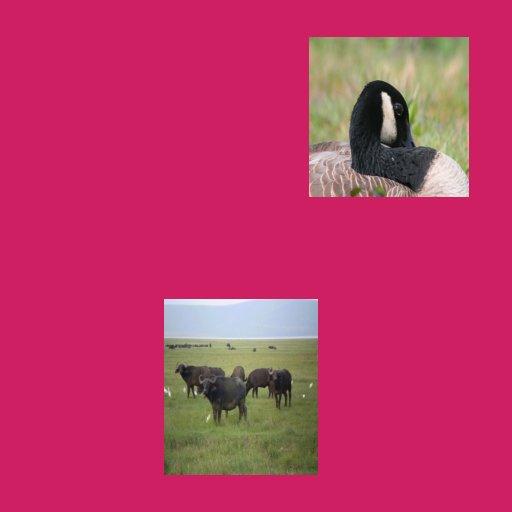}} &
\parbox[c]{6cm}{\raggedright\normalsize Q: Where is the goose located relative to the water buffalo? \\ Response format: The goose is located in the top-left/top-right/bottom-left/bottom-right of the water buffalo. \\[8pt]
A: The goose is located in the top-right of the water buffalo.} &
\raisebox{-0.5\height}{\includegraphics[width=5cm]{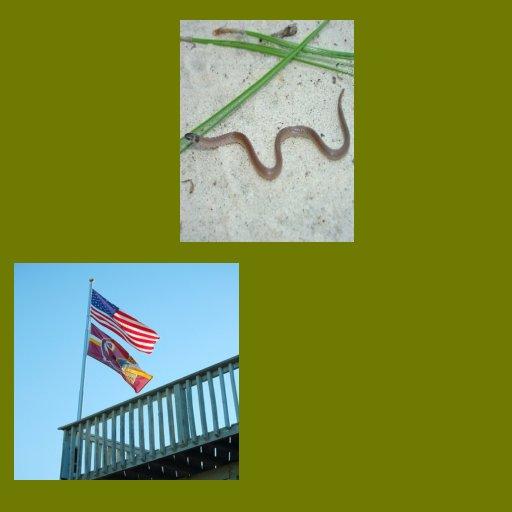}} &
\parbox[c]{6cm}{\raggedright\normalsize Q: If you look at the flagpole, where is the thunder snake? \\
Response format: The thunder snake is to the top-left/top-right/bottom-left/bottom-right of the flagpole. \\[8pt]
A: The thunder snake is to the top-right of the flagpole.} \\
\addlinespace[10pt]
\raisebox{-0.5\height}{\includegraphics[width=5cm]{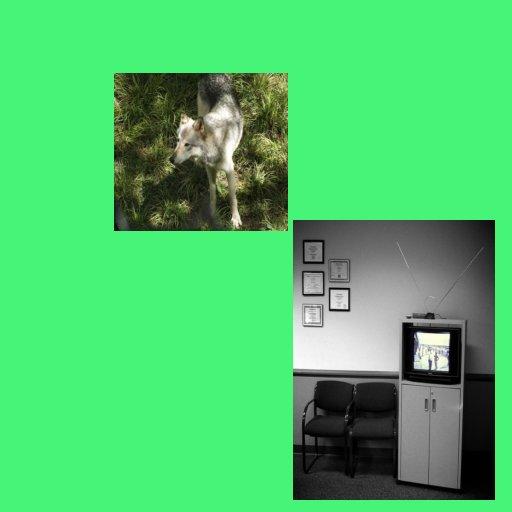}} &
\parbox[c]{6cm}{\raggedright\normalsize Q: Looking at the image, where does the timber wolf appear relative to the television? \\
Response format: The timber wolf appears to the top-left/top-right/bottom-left/bottom-right of the television. \\[8pt]
A: The timber wolf appears to the top-left of the television.} &
\raisebox{-0.5\height}{\includegraphics[width=5cm]{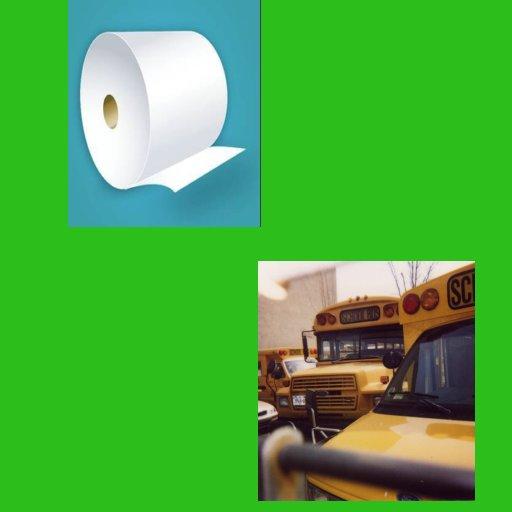}} &
\parbox[c]{6cm}{\raggedright\normalsize Q: How is the paper towel positioned with respect to the school bus? \\
Response format: The paper towel is at the top-left/top-right/bottom-left/bottom-right of the school bus. \\[8pt]
A: The paper towel is at the top-left of the school bus.} \\
\addlinespace[10pt]
\raisebox{-0.5\height}{\includegraphics[width=5cm]{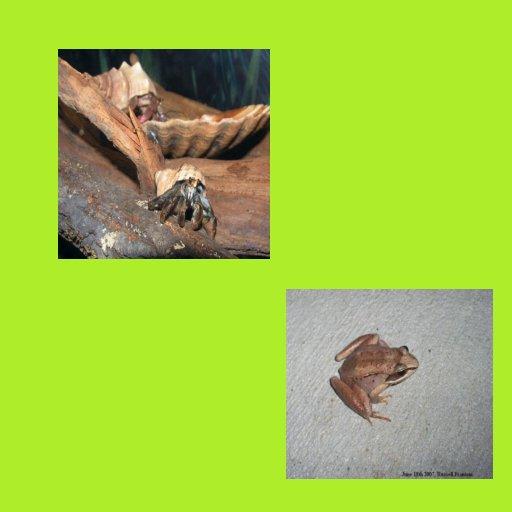}} &
\parbox[c]{6cm}{\raggedright\normalsize Q: Describe the spatial relationship between the tailed frog and the hermit crab. \\
Response format: The tailed frog is in the top-left/top-right/bottom-left/bottom-right relative to the hermit crab. \\[8pt]
A: The tailed frog is in the bottom-right relative to the hermit crab.} &
\raisebox{-0.5\height}{\includegraphics[width=5cm]{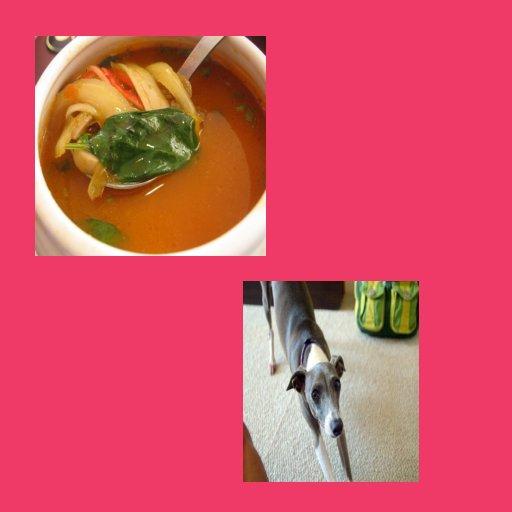}} &
\parbox[c]{6cm}{\raggedright\normalsize Q: How is the Italian greyhound positioned with respect to the soup bowl? \\
Response format: The Italian greyhound is at the top-left/top-right/bottom-left/bottom-right of the soup bowl. \\[8pt]
A: The Italian greyhound is at the bottom-right of the soup bowl.} \\
\addlinespace[10pt]
\raisebox{-0.5\height}{\includegraphics[width=5cm]{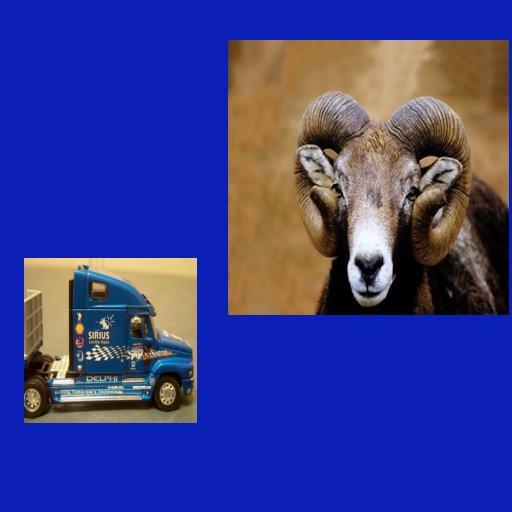}} &
\parbox[c]{6cm}{\raggedright\normalsize Q: Looking at the image, where does the trailer truck appear relative to the bighorn? \\
Response format: The trailer truck appears to the top-left/top-right/bottom-left/bottom-right of the bighorn. \\[8pt]
A: The trailer truck appears to the bottom-left of the bighorn.} &
\raisebox{-0.5\height}{\includegraphics[width=5cm]{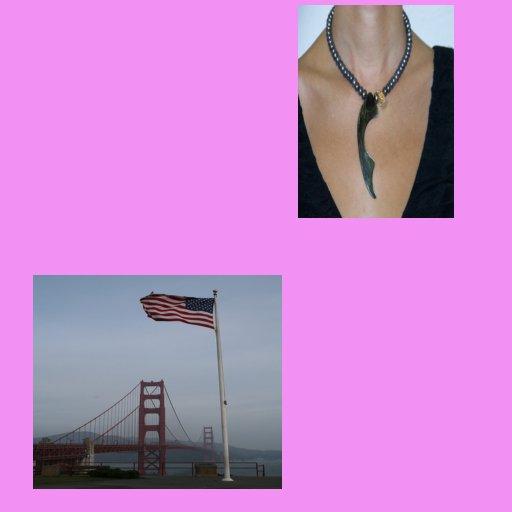}} &
\parbox[c]{6cm}{\raggedright\normalsize Q: What is the position of the flagpole relative to the necklace? \\
Response format: The flagpole is positioned to the top-left/top-right/bottom-left/bottom-right of the necklace. \\[8pt]
A: The flagpole is positioned to the bottom-left of the necklace.} \\
\addlinespace[4pt]
\bottomrule
\end{tabular}
}
\caption{Examples of spatial relation dataset.}
\label{tab:spatial_dataset}
\end{table*}

\subsubsection{Understanding evaluation.} 
We evaluate spatial relation understanding following the multiple-choice setting with 250 validation samples. Each original open-ended QA pair is
converted into a four-way multiple-choice question by removing the
\texttt{Response format} line and appending the following fixed choices:
\begin{quote}
\texttt{a. top-left}\\
\texttt{b. top-right}\\
\texttt{c. bottom-left}\\
\texttt{d. bottom-right}
\end{quote}
\noindent The ground-truth answer is mapped to the corresponding option
letter according to the target spatial relation.

\noindent An example evaluation pair after this transformation is:
\begin{quote}
\textbf{Question:} \\
\texttt{"Where is the goose located relative to the water buffalo?}\\
\texttt{a. top-left}\\
\texttt{b. top-right}\\
\texttt{c. bottom-left}\\
\texttt{d. bottom-right"}\\
\textbf{Answer:} \texttt{"b"}
\end{quote}

\subsubsection{Generation evaluation.} 
\begin{figure*}[t]
\centering
\fbox{%
\begin{minipage}{0.96\textwidth}
\vspace{6pt}
\tiny
\setlength{\fboxsep}{2pt}
\renewcommand{\baselinestretch}{0.9}
\noindent
\begin{minipage}[t]{0.48\textwidth}
\ttfamily
1. The dog located in the top-left of a teddy bear.\\
2. The wine glass located in the bottom-right of a kite.\\
3. The couch is located in bottom-left of a cup.\\
4. The laptop can be found to the top-right of a cow.\\
5. The fork is located in top-right of a hair drier.\\
6. The tie can be found to the bottom-left of a baseball bat.\\
7. The stop sign is located in top-right of a fork.\\
8. The bird is located in bottom-left of a skateboard.\\
9. The an apple located in the bottom-right of a tv.\\
10. The train located in the bottom-right of a potted plant.\\
11. The truck can be found to the top-right of a refrigerator.\\
12. The tv remote is located in bottom-right of a cow.\\
13. The bottle can be found to the bottom-left of a train.\\
14. The dog is located in top-right of a cow.\\
15. The skateboard located in the bottom-left of a person.\\
16. The baseball glove is located in bottom-left of an umbrella.\\
17. The dining table located in the top-right of an oven.\\
18. The hot dog can be found to the top-left of a suitcase.\\
19. The bus is located in top-left of a toothbrush.\\
20. The backpack located in the top-right of a sandwich.\\
21. The cake is located in bottom-right of a baseball bat.\\
22. The dog located in the top-left of a tie.\\
23. The suitcase can be found to the bottom-left of a boat.\\
24. The bear located in the bottom-right of a clock.\\
25. The tv remote can be found to the top-right of an umbrella.\\
26. The sports ball can be found to the top-left of an umbrella.\\
27. The train located in the top-right of a dining table.\\
28. The hair drier is located in bottom-right of an elephant.\\
29. The tennis racket can be found to the bottom-left of a spoon.\\
30. The wine glass can be found to the bottom-left of a hot dog.\\
31. The computer mouse can be found to the top-right of a bench.\\
32. The carrot can be found to the top-right of an orange.\\
33. The kite located in the bottom-right of a toothbrush.\\
34. The toaster is located in top-left of a traffic light.\\
35. The cat is located in top-left of a baseball glove.\\
36. The skis located in the top-left of a zebra.\\
37. The stop sign located in the bottom-left of a chair.\\
38. The stop sign located in the bottom-right of a parking meter.\\
39. The hot dog located in the top-left of a skateboard.\\
40. The pizza is located in bottom-left of a computer keyboard.\\
41. The hair drier can be found to the top-left of a toilet.\\
42. The cow can be found to the bottom-right of a stop sign.\\
43. The suitcase is located in top-right of a skis.\\
44. The book located in the bottom-left of a laptop.\\
45. The toothbrush is located in top-left of a pizza.\\
46. The toilet can be found to the top-right of a kite.\\
47. The tie is located in top-right of a sink.\\
48. The bird can be found to the bottom-right of a couch.\\
49. The bed can be found to the bottom-left of a sports ball.\\
50. The an elephant is located in top-left of a surfboard.\\
\end{minipage}
\hfill
\begin{minipage}[t]{0.48\textwidth}
\ttfamily
51. The frisbee can be found to the bottom-left of a motorcycle.\\
52. The vase located in the bottom-right of a fire hydrant.\\
53. The zebra can be found to the top-left of an elephant.\\
54. The bench can be found to the bottom-right of a bear.\\
55. The donut located in the top-left of a bench.\\
56. The frisbee is located in bottom-left of a horse.\\
57. The computer keyboard located in the bottom-right of a snowboard.\\
58. The tv is located in bottom-right of a cow.\\
59. The an elephant is located in top-left of a horse.\\
60. The suitcase can be found to the top-right of a banana.\\
61. The train is located in top-left of an airplane.\\
62. The cat is located in bottom-right of a backpack.\\
63. The backpack is located in bottom-left of a cake.\\
64. The sandwich is located in bottom-left of a knife.\\
65. The bicycle is located in top-right of a parking meter.\\
66. The knife located in the top-left of a suitcase.\\
67. The hot dog located in the bottom-left of a knife.\\
68. The zebra located in the top-left of a parking meter.\\
69. The chair can be found to the bottom-right of a zebra.\\
70. The cow is located in top-left of an airplane.\\
71. The cup can be found to the bottom-right of an umbrella.\\
72. The zebra is located in bottom-left of a computer keyboard.\\
73. The zebra is located in bottom-left of a broccoli.\\
74. The laptop is located in bottom-right of a sports ball.\\
75. The truck can be found to the top-left of a baseball bat.\\
76. The refrigerator is located in top-right of a baseball bat.\\
77. The tv located in the bottom-right of a baseball bat.\\
78. The baseball glove located in the top-left of a bear.\\
79. The refrigerator is located in bottom-right of a scissors.\\
80. The dining table located in the bottom-right of a suitcase.\\
81. The parking meter located in the bottom-right of a broccoli.\\
82. The frisbee located in the bottom-right of a truck.\\
83. The pizza located in the top-left of a banana.\\
84. The bus located in the bottom-left of a boat.\\
85. The cell phone can be found to the bottom-right of a tennis racket.\\
86. The horse located in the top-right of a broccoli.\\
87. The broccoli located in the bottom-left of a bottle.\\
88. The vase can be found to the bottom-left of a horse.\\
89. The bear is located in top-right of a spoon.\\
90. The zebra located in the top-left of a bed.\\
91. The cow can be found to the bottom-left of a laptop.\\
92. The bed located in the top-left of a frisbee.\\
93. The tie located in the top-right of a motorcycle.\\
94. The laptop can be found to the bottom-left of a tv.\\
95. The cell phone can be found to the bottom-left of a chair.\\
96. The couch is located in top-left of a potted plant.\\
97. The clock is located in bottom-left of a tv.\\
98. The couch is located in bottom-right of a vase.\\
99. The donut is located in bottom-left of a cat.\\
100. The couch can be found to the top-left of a toaster.\\
\end{minipage}
\vspace{6pt}
\end{minipage}%
}
\caption{\textbf{Full list of spatial relation evaluation prompts.}
All 100 prompts used for spatial relation evaluation.
Each prompt describes the relative position of two COCO objects using one of the predefined position templates.}
\label{fig:spatial_prompts}
\end{figure*}
We generate 100 prompts describing spatial relationships between two objects, following the same answer-style declarative format used during training. Each prompt is a sentence that states the diagonal relation between two object classes, \eg, ``The truck can be found to the top-left of the baseball.'' The full list of evaluation prompts is provided in Fig.~\ref{fig:spatial_prompts}. For each prompt, we generate four images with different random seeds and report the average accuracy across all generated samples.

For evaluation, we follow the same detection-based approach as GenEval~\cite{ghosh2023geneval}, obtaining bounding boxes for both objects. 
The relative position between objects is determined by the difference 
between bounding box centroids: if object $A$ is centered at $(x_A, y_A)$ 
and object $B$ is centered at $(x_B, y_B)$, following standard image 
coordinate conventions where the $y$-axis increases downward, then:
\begin{itemize}[label=\textbullet]
    \item \textbf{top-left:} $x_A < x_B$ and $y_A < y_B$
    \item \textbf{top-right:} $x_A > x_B$ and $y_A < y_B$
    \item \textbf{bottom-left:} $x_A < x_B$ and $y_A > y_B$
    \item \textbf{bottom-right:} $x_A > x_B$ and $y_A > y_B$
\end{itemize}
A sample is counted as correct if the detected relation matches the one 
specified in the prompt. If the detector fails to localize either object, 
the sample is excluded from evaluation.

\subsection{Text Recognition and Generation}
\subsubsection{Dataset details.} We construct the synthetic OCR training set from a public English sentence corpus on Hugging Face. Specifically, we use the training split of \path{agentlans/high-quality-english-sentences} and keep the first 200K samples. Each sample is normalized by replacing line breaks with spaces, trimming whitespace, and truncating the target string to at most 120 characters. The normalized text is then formatted as \texttt{\# <sentence>}, \ie, a single level-1 Markdown heading rather than a full free-form document. We convert this markdown string to HTML using the Python \texttt{markdown} library with standard extensions for fenced code blocks, tables, automatic line breaks, and list handling, and then rasterize the HTML into a PNG image with \texttt{wkhtmltoimage}. We define four hand-designed CSS templates, \texttt{clean\_light}, \texttt{dark}, \texttt{sepia}, and \texttt{blue\_gray}, which vary the background, text color, and font family. During understanding training, one of these templates is sampled at random for each example, whereas generation training uses the \texttt{clean\_light} template throughout. Figure~\ref{fig:ocr_dataset} shows representative examples.

We use the same underlying text-image pair for both modalities. For understanding, each rendered image is paired with the fixed instruction \texttt{Extract all text from the image.}, and the normalized sentence is used as the target. For generation, the same sentence is embedded into a fixed caption template describing a Mathpix-style rendering with sharp, legible text in natural reading order. This construction lets us compare \textit{und} and \textit{gen} transfer on identical textual content while changing only the direction of supervision. The explicit understanding and generation formats are shown below.

\begin{quote}
\textbf{Understanding question:} \texttt{"Extract all text from the image."} \\
\textbf{Understanding answer:} \texttt{"\{SENTENCE\}"} \\
\textbf{Generation prompt:} \texttt{"A Mathpix Markdown format with sharp, legible black text. High-resolution typography, top-down view. The text is rendered in natural left-to-right, top-to-bottom reading order. The text reads: \{SENTENCE\}"} \\
\end{quote}

\subsubsection{Understanding evaluation.} We evaluate text recognition on the first 250 samples from the test split of the same corpus. For each example, we normalize the reference sentence in the same way as during training and re-render it as a \texttt{clean\_light} markdown image. At evaluation time, we use the explicit instruction \texttt{Extract all text from the image in reading order.} and otherwise follow the default understanding prompt format and decoding procedure of each model. After generation, both prediction and reference are lowercased, line breaks are replaced with spaces, and leading and trailing whitespace is stripped. We report WER, CER, BLEU, METEOR, average edit distance, and token-level precision, recall, and F1.

\subsubsection{Generation evaluation.} We evaluate text generation on the first 250 samples from the test split. For each target sentence, we construct the same Mathpix-style caption template shown above and generate an image following the default text-to-image inference setting of each model. We then apply GLM-OCR~\cite{glm_ocr} to the generated image and extract text. The extracted text and the ground-truth sentence are normalized with the same lowercasing and whitespace processing used in understanding evaluation, and we report the same set of metrics: WER, CER, BLEU, METEOR, average edit distance, and token-level precision, recall, and F1.

\clearpage


\section{Additional Results and Analysis}

\subsection{Additional results on shared transformer with unified image encoder using MMaDA~\cite{yang2025mmada}}

\begin{table}[!h]
\centering
\setlength{\tabcolsep}{4pt}
\vspace{-2mm}

\begin{tabular}{c|cc|c}

\toprule

\multirow{2}{*}{\textbf{Train $\rightarrow$ Test}}
& \multicolumn{2}{c|}{\textbf{Counting}}
& \textbf{Spatial Rel.} \\

& Acc. (\%)~$\uparrow$
& MAD~$\downarrow$
& Acc. (\%)~$\uparrow$ \\

\hline

\rowcolor{gray!15}
\multicolumn{4}{l}{\textit{Understanding Evaluation}} \\

Baseline
& 38.7
& 1.16
& 40.8 \\

\genund
& 43.9 \good{+5.2}
& 1.07 \good{-0.09}
& 48.8 \good{+8.0} \\

\hline

\rowcolor{gray!15}
\multicolumn{4}{l}{\textit{Generation Evaluation}} \\

Baseline
& 15.8
& 3.44
& 33.0 \\

\undgen
& 20.0 \good{+4.2}
& 3.23 \good{-0.21}
& 46.3 \good{+13.3} \\

\bottomrule
\end{tabular}

\vspace{5pt}

\caption{\textbf{Bidirectional transferability of MMaDA~\cite{yang2025mmada}.}
  MMaDA shares the same architecture and generation objective as Lumina-DiMOO (Fig.~\ref{fig:arch_categorize}-(a), a shared transformer with a unified image encoder). We report counting (accuracy and MAD) and spatial-relation (accuracy) results under both \undgen\ transfer and \genund\ transfer. Consistent with Lumina-DiMOO, MMaDA improves on both tasks in both directions, indicating that transferability is shared across models of this architectural family.}
\label{tab:mmada}

\vspace{-20pt}
\end{table}

Due to computational cost and limited public code availability, our main experiments select one representative model from each architectural family in Fig.~\ref{fig:arch_categorize}. Among them, Lumina-DiMOO, which belongs to the shared transformer with a unified image encoder family (Fig.~\ref{fig:arch_categorize}-(a)), exhibits the strongest cross-task transferability, so we center our analysis on it in Secs.~\ref{sec:tasks} and~\ref{sec:discussion}. To verify that this behavior is not specific to a single model, we additionally evaluate MMaDA~\cite{yang2025mmada}, which shares the same architecture (Fig.~\ref{fig:arch_categorize}-(a)) and generation objective but differs in training data and scale.

As shown in Table~\ref{tab:mmada}, MMaDA exhibits transferability in both directions, mirroring the trend observed for Lumina-DiMOO. On the understanding side, \genund\ transfer improves counting accuracy from $38.7\%$ to $43.9\%$ ($+5.2$) and reduces MAD from $1.16$ to $1.07$ ($-0.09$), while spatial-relation accuracy rises from $40.8\%$ to $48.8\%$ ($+8.0$). On the generation side, \undgen\ transfer raises counting accuracy from $15.8\%$ to $20.0\%$ ($+4.2$) with MAD dropping from $3.44$ to $3.23$ ($-0.21$), and spatial-relation accuracy improves substantially from $33.0\%$ to $46.3\%$ ($+13.3$). The gains are consistent across both tasks and both transfer directions, which is in line with cross-task knowledge sharing rather than task-specific artifacts.

Although empirical, these results offer additional support for our main observation. A second, independently trained model from the shared transformer with a unified image encoder family (Fig.~\ref{fig:arch_categorize}-(a)) again exhibits bidirectional transferability, consistent with the trend in Table~\ref{tab:transfer}. We read this as further evidence that transferability is associated with this architectural design rather than with any individual model. We note, however, that a stronger claim would require broader coverage across more models and families, which remains constrained by computational cost and limited public code availability.

\subsection{Attention-level Analysis of Transferability}

\begin{figure}[!h]
    \vspace*{\fill}
    \includegraphics[width=\linewidth]{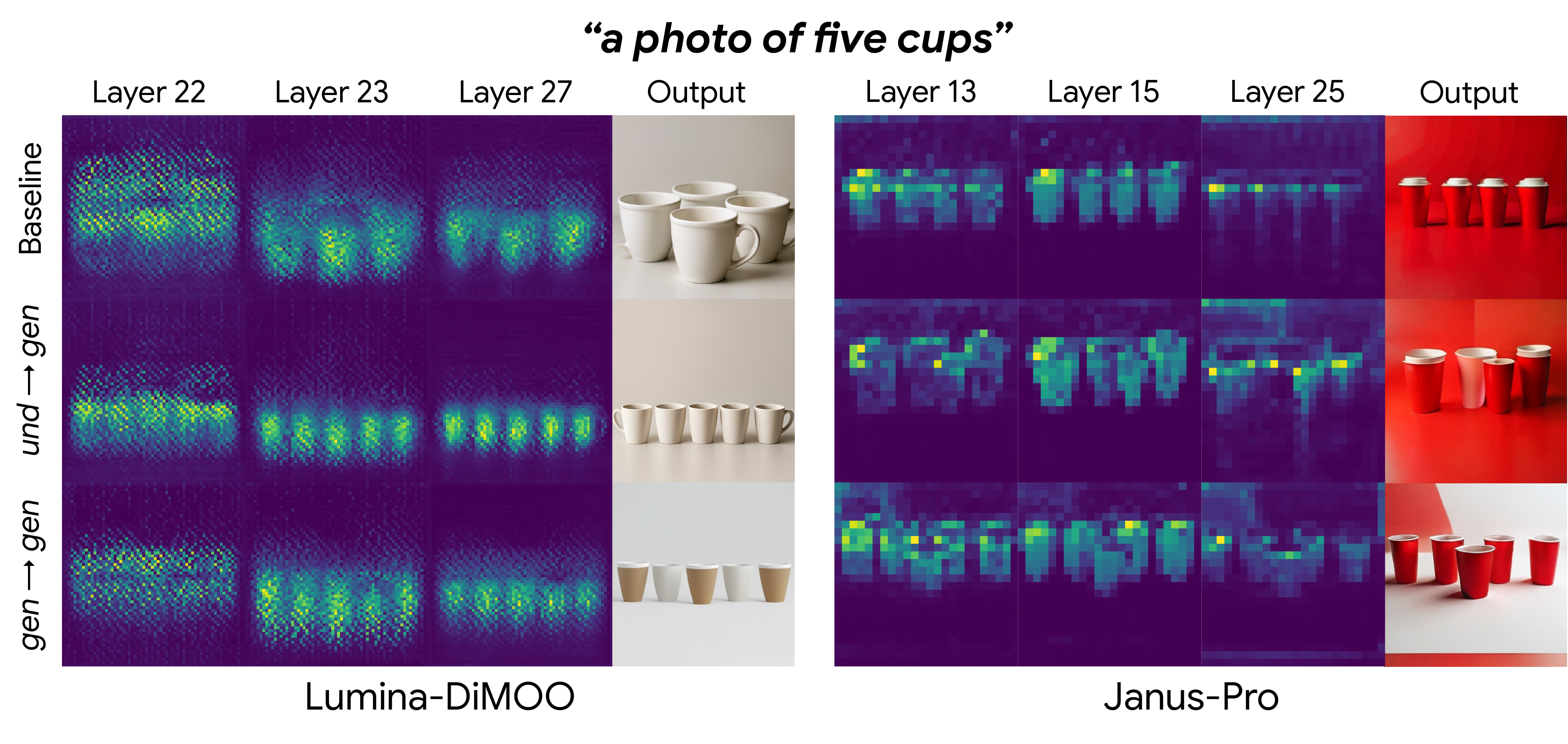}
    \vspace{-10pt}
    \caption{\textbf{Attention analysis on Lumina-DiMOO and Janus-Pro on counting.} 
    We visualize the attention maps associated with the text token ``cups'' in the prompt ``a photo of five cups''. The attention scores are averaged over heads and reshaped into the spatial layout of image tokens. In Lumina-DiMOO, which exhibits strong transferability, \undgen\ (middle row) produces sharper and more prompt-aligned attention over cup regions than the baseline (top row), to a degree comparable to direct \gengen\ (bottom row). In Janus-Pro, where such transfer is absent, the attention maps remain largely unchanged across training settings.
    }
    \label{fig:attention}
    \vspace{-10pt}
\end{figure}

Beyond the output-level evaluation, we further inspect the intermediate behavior of Lumina-DiMOO~\cite{xin2025lumina} and Janus-Pro~\cite{chen2025januspro} by qualitatively analyzing their multimodal attention maps on the counting task~(Fig.~\ref{fig:attention}).Specifically, we visualize the text-to-image attention scores in the multimodal attention layers, where the query corresponds to a text token and the keys correspond to image tokens involved in generation.  Since the two models follow different generation paradigms, we extract attention maps at model-specific generation stages. For Lumina-DiMOO, a discrete diffusion model, we cache the multimodal attention maps at each denoising step and visualize the attention scores at an early denoising stage (after two denoising steps out of total 64 denoising steps) when the coarse image structure begins to emerge. For Janus-Pro, an autoregressive image generation model, we visualize the attention scores after the image generation process is completed. 

The resulting attention patterns reveal a qualitative difference that aligns with the transferability trends observed in Table~\ref{tab:transfer}. In Lumina-DiMOO, where understanding-to-generation transfer is strong, the \undgen\ model produces attention that is not only more localized over the generated cup regions compared with the baseline, but also more consistent with the count-object relationship specified in the prompt. This pattern appears across several layers and is visually comparable to that of direct generation training (\gengen). In contrast, Janus-Pro, which does not show clear understanding-to-generation transfer, exhibits much weaker changes in text-to-image attention across the baseline, \undgen\, and \gengen\ models. These observations suggest that, when transferability emerges, understanding-side training may also affect how textual concepts are routed to image tokens during generation.


\clearpage
\subsection{Quantitative Analysis of Generation Distribution Shift After Fine-Tuning}

\label{sec:appendix_gen_results}
\begin{table*}[!]
\centering
\begin{minipage}[t]{0.49\textwidth}
\centering
\textbf{Task-Specific Prompts}\par\vspace{4pt}
\resizebox{\linewidth}{!}{%
\begin{tabular}{l|cc|cc|cc}
\toprule
\multirow{2}{*}{\textbf{Model}}
& \multicolumn{2}{c|}{\textbf{Baseline}}
& \multicolumn{2}{c|}{\textbf{\undgen}}
& \multicolumn{2}{c}{\textbf{\gengen}} \\
& \textbf{IS}~$\uparrow$ & \textbf{FID}~$\downarrow$
& \textbf{IS}~$\uparrow$ & \textbf{FID}~$\downarrow$
& \textbf{IS}~$\uparrow$ & \textbf{FID}~$\downarrow$ \\
\midrule
\rowcolor{gray!15}
\multicolumn{7}{l}{\textit{Counting}} \\
Bagel & 10.37 & - & 10.39 & 9.03 & 10.80 & 12.85 \\
Janus-Pro & 8.45 & - & 8.61 & 13.69 & 9.23 & 43.00 \\
Lumina-DiMOO & 7.93 & - & 7.88 & 14.74 & 9.20 & 80.45 \\
BLIP3-o & 7.69 & - & 7.90 & 15.66 & 8.06 & 85.76 \\
\midrule
\rowcolor{gray!15}
\multicolumn{7}{l}{\textit{Spatial Relation}} \\
Lumina-DiMOO & 16.25 & - & 16.28 & 13.26 & 15.30 & 24.90 \\
\midrule
\rowcolor{gray!15}
\multicolumn{7}{l}{\textit{Text Recognition / Generation}} \\
Lumina-DiMOO & 1.14 & - & 1.14 & 4.29 & 1.10 & 21.45 \\
\bottomrule
\end{tabular}
}
\end{minipage}\hfill
\begin{minipage}[t]{0.49\textwidth}
\centering
\textbf{COCO Validation Prompts}\par\vspace{4pt}
\resizebox{\linewidth}{!}{%
\begin{tabular}{l|cc|cc|cc}
\toprule
\multirow{2}{*}{\textbf{Model}}
& \multicolumn{2}{c|}{\textbf{Baseline}}
& \multicolumn{2}{c|}{\textbf{\undgen}}
& \multicolumn{2}{c}{\textbf{\gengen}} \\
& \textbf{IS}~$\uparrow$ & \textbf{FID}~$\downarrow$
& \textbf{IS}~$\uparrow$ & \textbf{FID}~$\downarrow$
& \textbf{IS}~$\uparrow$ & \textbf{FID}~$\downarrow$ \\
\midrule
\rowcolor{gray!15}
\multicolumn{7}{l}{\textit{Counting}} \\
Bagel & 19.32 & - & 18.94 & 15.91 & 18.97 & 21.89 \\
Janus-Pro & 17.31 & - & 18.74 & 37.10 & 17.72 & 43.89 \\
Lumina-DiMOO & 16.86 & - & 17.55 & 31.47 & 15.29 & 52.51 \\
BLIP3-o & 17.84 & - & 18.61 & 29.17 & 15.69 & 59.40 \\
\midrule
\rowcolor{gray!15}
\multicolumn{7}{l}{\textit{Spatial Relation}} \\
Lumina-DiMOO & 16.86 & - & 17.50 & 30.96 & 17.41 & 32.28 \\
\midrule
\rowcolor{gray!15}
\multicolumn{7}{l}{\textit{Text Recognition / Generation}} \\
Lumina-DiMOO & 16.86 & - & 18.03 & 30.75 & 17.51 & 31.19 \\
\bottomrule
\end{tabular}
}
\end{minipage}

\vspace{5pt}
\caption{\textbf{Generation quality across unified multimodal models.} Left: task-specific prompts. Right: COCO validation prompts. We report Inception Score (IS) and Fr\'echet Inception Distance (FID) for the baseline, \undgen, and \gengen checkpoints. Since FID is computed with images generated by baseline, baseline FID is undefined and is therefore shown as `-'.}
\label{tab:appendix_fid_is_combined}

\vspace{-10pt}
\end{table*}


We measure FID~\cite{heusel2017fid} and IS~\cite{salimans2016isscore} under two prompt settings to assess how generation-side fine-tuning affects the output distribution of the baseline image generation. First, we evaluate with task-specific prompts that match the format used during generation training. We use 1.8K samples for counting, 2K for spatial relation, and 1.8K for text recognition and generation. Second, to examine the effect on more general image synthesis, we repeat the evaluation using 1K prompts from the COCO~\cite{lin2014coco} validation set. The results are reported in Table~\ref{tab:appendix_fid_is_combined}.

We first consider FID to quantify the distribution shift, which is computed with images generated by the baseline. Across both prompt settings and all tasks, \undgen\ consistently yields lower FID than \gengen. This result indicates that understanding-based tuning preserves the baseline's original output image distribution more faithfully than direct generation tuning.

By contrast, IS is considerably more task-dependent. Under the task-specific prompt setting, the IS of \undgen\ is generally close to that of the baseline, whereas \gengen\ often attains higher IS on counting. One possible explanation is that direct generation tuning on the counting dataset, which contains multiple instances of the same object category, makes object appearance more classifier-friendly within the narrow counting prompt distribution. However, this tendency does not extend consistently to spatial relation or text recognition and generation, and under the COCO~\cite{lin2014coco} validation prompt set the apparent advantage of \gengen\ weakens or disappears. We therefore interpret changes in IS not as direct evidence of improved general image quality, but rather as reflecting adaptation to a narrow task-specific prompt distribution. 

Overall, these results suggest that \undgen\ perturbs the baseline image-generation distribution substantially less than \gengen.

\clearpage
\subsection{Ablation on how LoRA rank affects transfer strength}

\begin{table}[h]
\centering
\setlength{\tabcolsep}{4pt}
\vspace{-2mm}

\begin{tabular}{c|cc}

\toprule

& \multicolumn{2}{c}{Counting} \\
LoRA Rank & Acc.(\%)~$\uparrow$ & MAD~$\downarrow$ \\

\hline

\rowcolor{gray!15}
\multicolumn{3}{l}{\textit{Generation Evaluation}} \\

Baseline & 48.0 & 1.11  \\

8
& 51.9 \good{+3.9}
& 1.01 \good{-0.10}  \\

32
& 53.0 \good{+5.0}
& 0.95 \good{-0.16}  \\

128
& 53.6 \good{\textbf{+5.6}}
& 0.88 \good{\textbf{-0.23}} \\

\bottomrule
\end{tabular}

\vspace{5pt}
\caption{\textbf{Ablation on how LoRA rank affects \undgen\ transfer strength.}
  We evaluate counting \undgen\ transfer on Lumina-DiMOO across LoRA ranks 8, 32, and 128. Both counting accuracy and MAD improve monotonically with the rank, indicating that larger ranks yield stronger transfer.}
\label{tab:lora_rank_abl}

\vspace{-10pt}
\end{table}

A natural question follows from the observation that transfer emerges through training: does stronger training induce stronger transfer? To examine whether the extent of parameter adaptation correlates with the strength of transfer, we conduct an ablation over the LoRA rank, which controls how much the model parameters are allowed to change during fine-tuning. We compare counting \undgen\ transfer on Lumina-DiMOO across ranks 8, 32, and 128 at equal 3K training steps. As shown in Table~\ref{tab:lora_rank_abl}, the transfer strength grows monotonically with the rank: counting accuracy rises from $48.0\%$ at the baseline to $51.9\%$, $53.0\%$, and $53.6\%$ at ranks 8, 32, and 128, with MAD dropping from $1.11$ to $0.88$.

We also conducted full fine-tuning experiments. However, when we trained the full model on the counting dataset alone, training was much more unstable than LoRA (rank 128), and the resulting generation quality collapsed entirely. 

Overall, these results indicate that although stronger training generally leads to stronger transfer, properly exploiting transferability also requires choosing an appropriate tuning strength. We leave a systematic exploration of this trade-off to future work.

\clearpage
\subsection{Can transferred capabilities be naturally combined at generation time to enable more complex generation?}

\begin{table}[h]
\centering
\setlength{\tabcolsep}{4pt}

\begin{tabular}{c|c}

\toprule
Train $\rightarrow$ Test & \textbf{Joint Acc.(\%)~$\uparrow$}\\
\hline

\rowcolor{gray!15}
\multicolumn{2}{l}{\textit{Generation Evaluation}} \\

Baseline & 24.7 \\

\undgen
& 28.5 \good{+3.8} \\

\bottomrule
\end{tabular}

\vspace{10pt}
\caption{\textbf{Composing two transferred capabilities at generation time.}
  After fine-tuning Lumina-DiMOO on simple mixture of counting and spatial-relation \emph{understanding} data, we evaluate on 400 complex generation prompts requiring \textbf{\textit{both}} counting and spatial reasoning, e.g., \textit{``three cats are on the top-right of a sofa''}, which is an unseen scenario during training. Joint accuracy measures the fraction of target objects that satisfy both (1) the required object count and (2) the specified spatial relation. Trained model \textit{naturally composes} the two transferred capabilities enabling \textit{more complex generation behaviors}, despite never seeing such compositional training examples during training.}

\label{tab:composition}

\vspace{-10pt}

\end{table}

Our main experiments transfer a single capability at a time. A natural follow-up question is what happens when several capabilities are transferred together: do they remain isolated, each improving only its own attribute independently, or can they be \emph{combined} at generation time? To probe this, we train a single model on a mixture of counting and spatial-relation \emph{understanding} data, without any example that requires both capabilities jointly. We then evaluate it on 400 compositional generation prompts that demand the two capabilities simultaneously, \eg, \textit{``three cats are on the top-right of a sofa''}, which is an unseen scenario during training. Following our main protocol, we report joint accuracy, defined as the fraction of target objects that satisfy \textit{both} (1) the required object count and (2) the specified spatial relation.

As shown in Table~\ref{tab:composition}, the model improves joint accuracy from $24.7\%$ to $28.5\%$ ($+3.8\%$) over the baseline, even though it was never trained on compositional examples. This indicates that the capabilities transferred through understanding are not merely injected in isolation but can be composed during generation. We view this as preliminary evidence that more complex generation behaviors may be assembled from simpler component capabilities via \undgen\ transfer.

\clearpage
\section{Additional Qualitative Results}
\begin{figure}[!]
    \vspace*{\fill}
    \includegraphics[width=\linewidth]{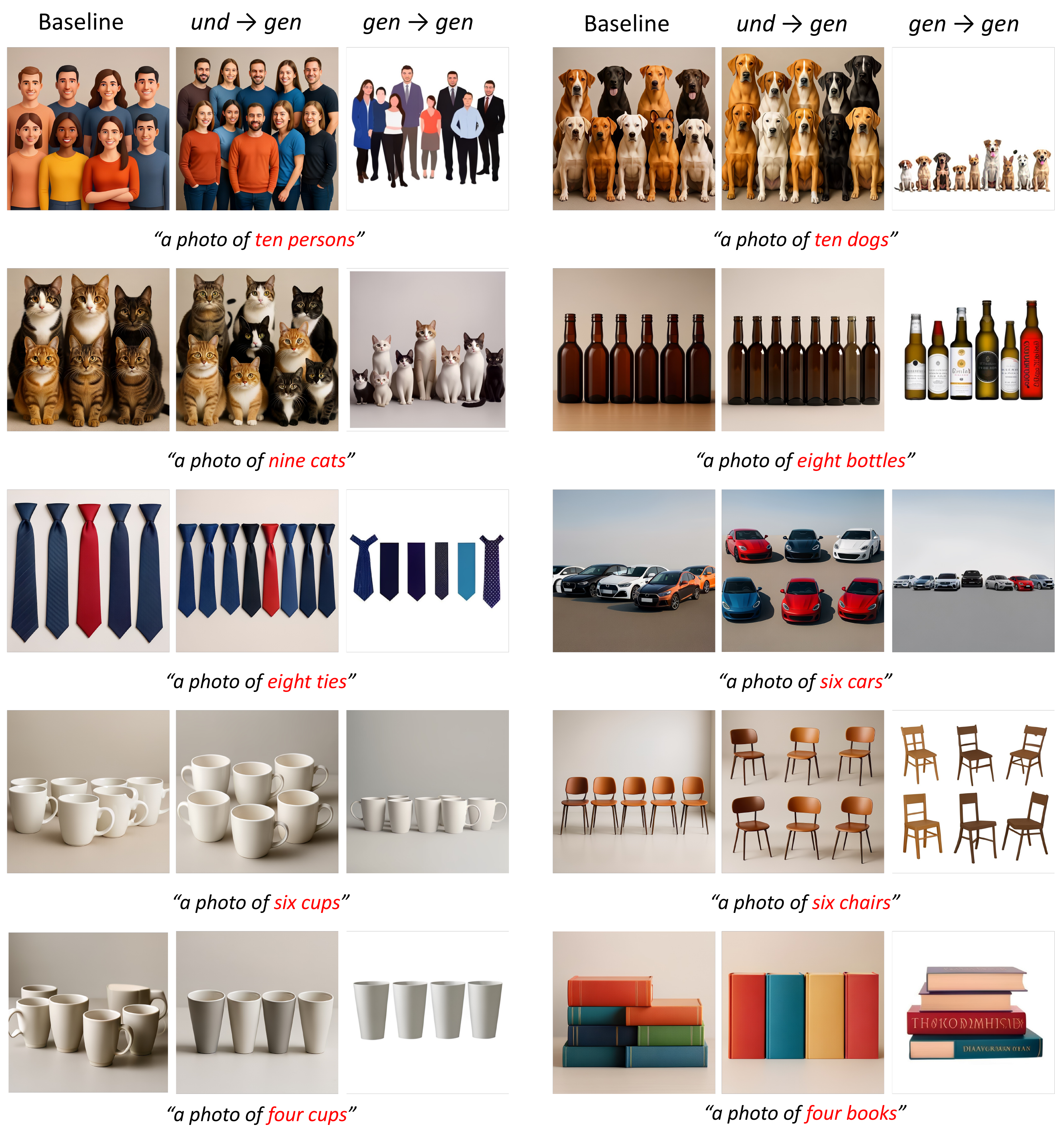}
    \vspace{-10pt}
    \caption{\textbf{Additional qualitative comparisons for Lumina-DiMOO between \undgen\ 
and \gengen on counting tasks.} }
    \label{fig:appendix_counting}
    \vspace{-10pt}
\end{figure}

\begin{figure}[p]
    \vspace*{\fill}
    \centering
    \includegraphics[width=\linewidth]{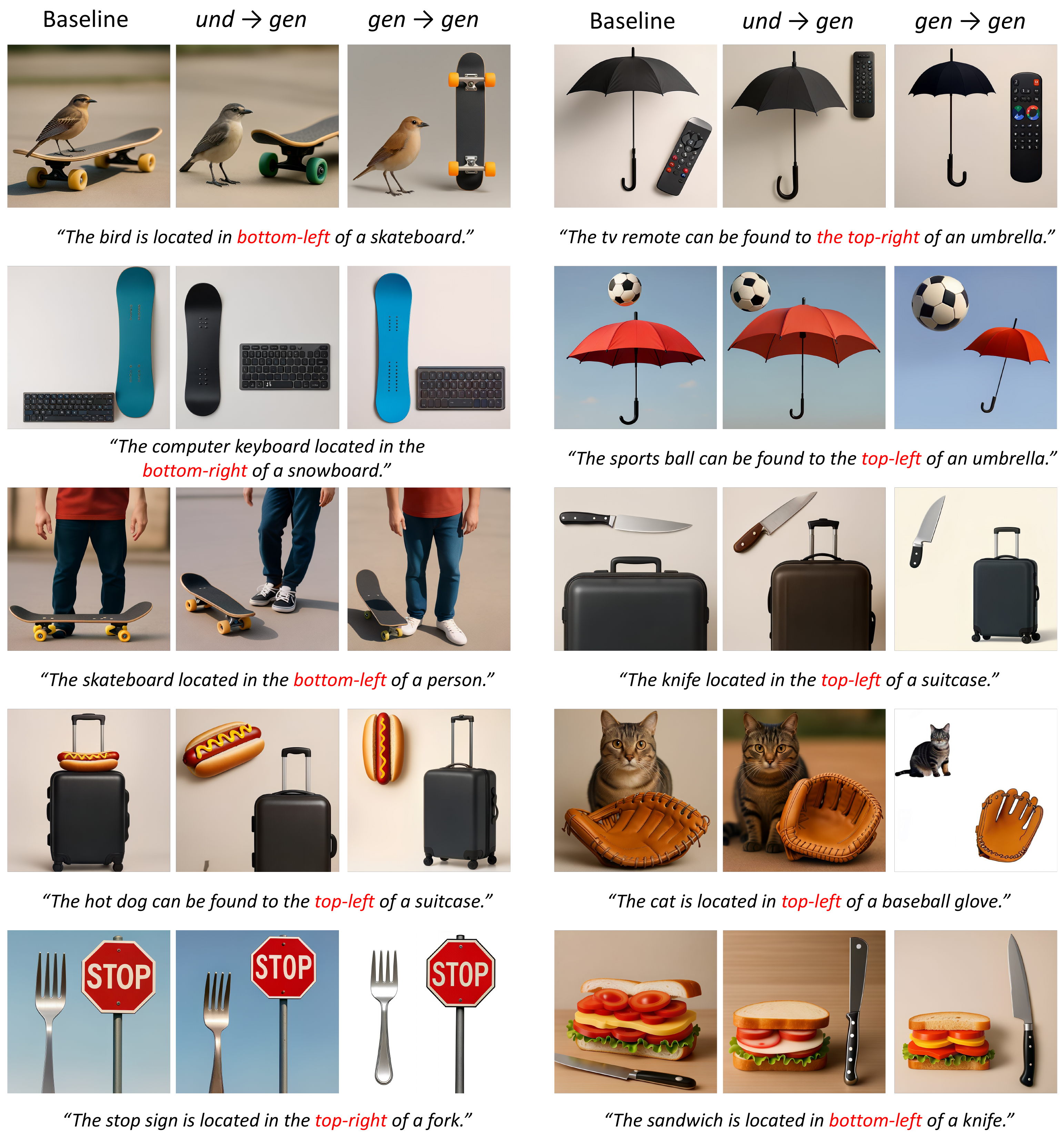}
    \vspace{-10pt}
    \caption{\textbf{Additional qualitative comparisons for Lumina-DiMOO between \undgen\ 
and \gengen on spatial relation tasks.} }
    \label{fig:appendix_spatial}
    \vspace{-10pt}
\end{figure}

\begin{figure}[p]
    \vspace*{\fill}
    \includegraphics[width=\linewidth]{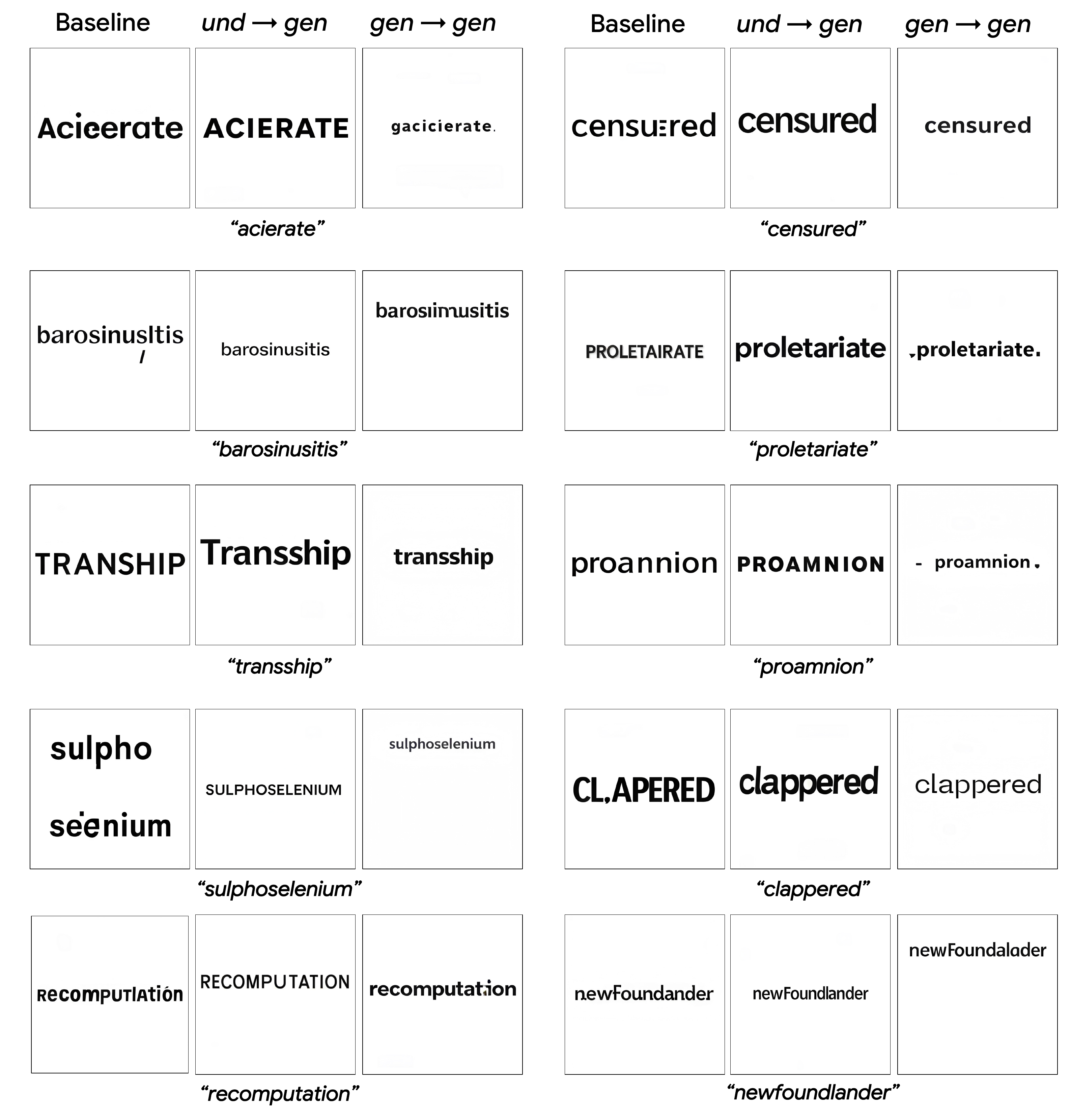}
    \vspace{-10pt}
    \caption{\textbf{Additional qualitative comparisons for Lumina-DiMOO between \undgen\ 
and \gengen on text generation tasks.} }
    \label{fig:appendix_textgen}
    \vspace{-10pt}
\end{figure}

We provide additional qualitative examples for Lumina-DiMOO in
Figures~\ref{fig:appendix_counting}, \ref{fig:appendix_spatial}, and \ref{fig:appendix_textgen}.
These figures extend the Lumina-DiMOO results shown in Figures~\ref{fig:undgen_counting},
\ref{fig:spatial_pos}, and \ref{fig:ocr_results} in the main paper.
\label{sec:appendix_qualitative}

\clearpage

\clearpage

\clearpage
\section{Limitations}
\label{sec:appendix_limitations}
In this study, we examine four representative UMMs drawn from different architectural families. While this selection allows us to explore how transferability varies with architectural design, it does not cover the full spectrum of possible UMM configurations. Evaluating a wider range of models would help assess how broadly our observations hold.

We also find that transferability cannot be attributed to architectural factors alone. Both its direction and magnitude depend on the type of visual knowledge required by each task. Although we present empirical results and qualitative insights highlighting this task dependence, determining the optimal transfer direction in advance remains challenging.

Future work could expand the analysis to more diverse model families and, importantly, investigate principled ways to anticipate transferability across tasks.

\end{document}